\documentclass{article}

\PassOptionsToPackage{dvipsnames}{xcolor}

\usepackage{iclr2027_conference,times}
\iclrfinalcopy

\usepackage[utf8]{inputenc} 
\usepackage[T1]{fontenc}    
\usepackage{hyperref}       
\usepackage{url}            
\usepackage{booktabs}       
\usepackage{amsfonts}       
\usepackage{nicefrac}       
\usepackage{microtype}      
\usepackage{xcolor}         
\usepackage{kotex}
\usepackage[textsize=tiny]{todonotes}
\usepackage{tabularx} 
\usepackage{booktabs}
\usepackage{graphicx}
\usepackage{kotex}
\usepackage{graphicx}
\usepackage{multirow}
\usepackage{colortbl}
\usepackage{makecell}
\usepackage{pifont}
\usepackage[normalem]{ulem}
\usepackage[all]{hypcap}
\usepackage{caption}
\usepackage{tcolorbox}
\tcbuselibrary{listings,breakable,skins}
\usepackage{arydshln}         
\usepackage{caption}
\usepackage{enumitem}
\usepackage{fancyvrb}
\usepackage{microtype}
\usepackage{listings}
\usepackage{inconsolata} 
\usepackage{wrapfig}
\usepackage{array}
\usepackage{amsmath} 
\usepackage{amssymb} 

\hypersetup{
  colorlinks=true,
  linkcolor={NavyBlue},
  citecolor={NavyBlue},
  urlcolor={NavyBlue},
}

\newcolumntype{Y}{>{\raggedright\arraybackslash}X}
\definecolor{omniblue}{HTML}{547AB4}
\definecolor{omniorange}{HTML}{DF8077}

\usepackage{fontawesome5} 
\newtcolorbox{keyfinding}[1]{%
  enhanced,
  breakable,
  colback=white,
  colframe=teal!70!black,
  fonttitle=\bfseries\sffamily,
  coltitle=white,
  boxrule=0.8pt,
  arc=3pt,
  left=10pt, right=10pt, 
  top=8pt,          
  bottom=4pt,       
  before skip=8pt,  
  after skip=8pt,   
  shadow={1.5mm}{-0.8mm}{0mm}{black!10}, 
  borderline west={3pt}{0pt}{teal!70!black},
  attach boxed title to top left={xshift=8pt, yshift=-2.5mm}, 
  boxed title style={
    sharp corners,
    colback=teal!70!black,
    boxrule=0pt,
    rounded corners=north,
    arc=2pt,
  },
  title={\small\faLightbulb\ #1}
}

\lstdefinelanguage{json}{
  basicstyle=\ttfamily\small,
  showstringspaces=false,
  breaklines=true,
  breakatwhitespace=true,
  columns=fullflexible
}

\tcbset{
  mypromptbox/.style={
    enhanced,
    breakable,
    colback=blue!2,          
    colframe=blue!70!black,  
    boxrule=0.8pt,           
    arc=8pt,
    left=8pt,right=8pt,top=6pt,bottom=6pt,
    fonttitle=\bfseries\large,
    coltitle=white,
    colbacktitle=blue!80!black
  }
}

\newcolumntype{C}[1]{>{\centering\arraybackslash}p{#1}}

\definecolor{myHotPink}{HTML}{FF1493}

\title{Omni-Persona: Systematic Benchmarking and Improving Omnimodal Personalization}

\newcommand{\leftauthorline}[1]{%
  \makebox[\dimexpr\textwidth-2\tabcolsep\relax][l]{%
    \hspace*{-\tabcolsep}#1%
  }%
}

\author{%
\leftauthorline{\bfseries
  Yeongtak Oh$^1$, Dongwook Lee$^2$, Sangkwon Park$^1$,
  Heeseung Kim$^3$, Sungroh Yoon$^{1,2}$%
  \thanks{Corresponding author}} \\[0.25em]
\leftauthorline{$^1$Department of Electrical and Computer Engineering, Seoul National University} \\
\leftauthorline{$^2$Interdisciplinary Program in Artificial Intelligence, Seoul National University} \\
\leftauthorline{$^3$Department of Artificial Intelligence, University of Seoul} \\[0.25em]
\leftauthorline{$^{1,2}$\texttt{\{dualism9306, dwsmart32, tkdrnjs0621, sryoon\}@snu.ac.kr}} \\
\leftauthorline{$^{3}$\texttt{gmltmd789@uos.ac.kr}}
}

\begin{document}

\maketitle
\lhead{Preprint. Under review.}
\vspace{-1.2em}

\begin{abstract}
While multimodal large language models have advanced across text, image, and audio, personalization research has remained primarily vision-language, with unified omnimodal benchmarking that jointly covers text, image, and audio still limited, and lacking the methodological rigor to account for absent-persona scenarios or systematic grounding studies. We introduce Omni-Persona, the first comprehensive benchmark for omnimodal personalization. We formalize the task as cross-modal routing over the \emph{Persona Modality Graph}, encompassing 4 task groups and 18 fine-grained tasks across ${\sim}750$ items. To rigorously diagnose grounding behavior, we propose \emph{Calibrated Accuracy ($\mathrm{Cal}$)}, which jointly evaluates correct grounding and appropriate abstention, incorporating absent-persona queries within a unified evaluation framework. On our dedicated experiments, three diagnostic findings emerge: (i) recent open-weight models show a consistent audio-vs-visual grounding gap that RLVR partially narrows via dense rule-based supervision; (ii) recall score and parameter scale are incomplete diagnostics, since strong recall can coexist with absent-persona hallucination and larger models do not always achieve higher $\mathrm{Cal}$, exposing calibration as a separate evaluation axis; and (iii) SFT is limited by the scalability of ground-truth annotation. RLVR improves recall but not calibrated abstention under our reward design, leaving $\mathrm{Cal}$ at or below the base model. Omni-Persona thus serves as a diagnostic framework that surfaces the pitfalls of omnimodal personalization, guiding future post-training and reward design. Project page:~\href{https://github.com/oyt9306/Omni-Persona}{github.com/oyt9306/Omni-Persona}
\end{abstract}

\section{Introduction}
The landscape of large generative models has expanded rapidly toward \textit{omnimodal} systems capable of processing or even generating across text, image, and audio within a single model~\citep{qwen25omni,minicpmo,comanici2025gemini,gemma4,openai2024}. This convergence of modalities broadens the task scope that a single model can handle and moves the community closer to the vision of a personal AI assistant, one that can recognize a user's face and voice, recall their biographical context, and ground responses in individual identity.

Despite this momentum, multimodal personalization research has remained primarily focused on vision-language settings~\citep{yollava,rap,repic,covip}, leaving three key gaps that limit progress toward true omnimodal deployment. First, existing benchmarks have rarely provided unified coverage across all three modalities: while vision and text are well represented, systematic treatment of audio signals such as voice identity, emotional tone, and conversational context remains limited. Second, real-world retrieval is inherently noisy, often yielding contexts where the queried identity is completely \emph{absent}. Yet, personalization is typically evaluated under \textit{well-controlled settings}, such as explicit identity naming~\citep{myvlm,yollava,repic} or carefully designed caption-based distractors~\citep{covip}, that assume the target is always present. Consequently, these artificial setups and their recall-only protocols fail to expose this critical failure mode. Third, \textit{realistic personalization scenarios} (for example, identifying a person from a face image or voice clip and then answering a query about that individual) have not been systematically studied. Without a benchmark that addresses all three gaps, the community lacks a principled way to diagnose \emph{when} and \emph{how} current omnimodal models fail at personal grounding. While recent studies~\citep{atmbench, kim2025mmpb} each address important aspects of the multimodal personalization problem, substantial gaps remain in audio grounding, absent-persona coverage, and realistic evaluation.

To this end, we introduce \emph{Omni-Persona}, the first evaluation-only benchmark for omnimodal personalization, offering systematic cross-modal coverage with full support for \emph{audio} as a persona modality and \emph{absent-persona} cases. We formalize each user's multimodal profile through the \textit{Persona Modality Graph} (PMG). In this graph-based abstraction, individual user profiles (comprising a profile image, biographical text, and personal audio) act as context nodes. We frame omnimodal personalization as a cross-modal routing problem: the model must evaluate incoming queries and correctly establish a directed linkage (edge) to the matching context node to ground its response. 

Omni-Persona spans 4 task groups and 18 fine-grained tasks over ${\sim}750$ evaluation items, enabling evaluation of both perceptual matching and grounded retrieval. To reflect real-world retrieval imperfections, we include absent-persona samples, where the ground-truth persona is missing from the retrieved context. This setting introduces retrieval noise and captures a crucial challenge overlooked by prior multimodal personalization benchmarks. Finally, because recall alone cannot capture hallucination and over-abstention, we employ \emph{Calibrated accuracy} (Cal) as our primary metric, equally scoring correct grounding for answerable items and correct abstention for absent-persona items.

Beyond benchmarking, we investigate which post-training regimes best align current omnimodal models with personalization. While previous studies~\citep{repic, covip} have highlighted the efficacy of RLVR for multimodal personalization in image captioning tasks, we broaden this investigation to omnimodal personalization. Specifically, we rigorously compare supervised fine-tuning (SFT) and reinforcement learning with verifiable rewards (RLVR) to reveal which post-training regime is most suitable, and which specific aspects drive improvements in \emph{omnimodal personalization}. Prior work establishes that SFT is heavily influenced by data quality~\citep{zhou2023lima} and scale~\citep{dong2024abilities}, whereas recent RLVR methods rely on carefully specified verifiable reward signals, such as rule-based accuracy and format rewards~\citep{guo2025deepseek}. Motivated by this distinction, we conduct SFT on our rigorously curated ground-truth annotation corpora at two different scales (1K and 10K). We contrast this with an RLVR (without SFT warmup) recipe that jointly optimizes perception and retrieval. This RLVR approach utilizes rule-based perceptual verification, alongside LLM-as-a-judge retrieval verification for free-form QA.

Our comparative analysis reveals a clear trade-off. SFT is limited by the difficulty of constructing high-quality, in-domain ground-truth supervision for diverse open-ended scenarios; consequently, broader task coverage does not always translate into gains in $\mathrm{Cal}$. Our RLVR approach alleviates this limitation by using verifiable reward signals to optimize task-level correctness directly, without requiring reference responses for every training instance. However, it introduces a separate trade-off: under our reward design, models tend to favor answering over abstaining. We validate these findings comprehensively across both Qwen2.5-Omni and Gemma4~\citep{qwen25omni, gemma4} architectures.

Our contributions are as follows:
\begin{enumerate}[label=(\arabic*), leftmargin=*]
  \item \textbf{Omni-Persona Benchmark and PMG Formulation.}
  We introduce \emph{Omni-Persona}, the first comprehensive evaluation-only benchmark for omnimodal personalization. Built on the \emph{Persona Modality Graph} (PMG), it formalizes contextual grounding over retrieved persona evidence and the integration of raw-form multimodal contexts, spanning 4 task groups and 18 fine-grained tasks across image, text, and vocal audio.

  \item \textbf{Addressing Recall Blind Spots with Absent-Persona Evaluation.}
  While prior personalization benchmarks heavily rely on answerable-only recall, we elevate absent-persona queries to a first-class evaluation dimension. By coupling these unanswerable queries with hard distractors and retrieval noise, we propose a calibrated accuracy metric that jointly assesses correct grounding and appropriate abstention. This balanced approach exposes critical hallucination and over-abstention behaviors often masked by recall-only protocols.

    \item \textbf{Diagnostic Analysis of Omnimodal Personalization and Post-Training.}
    We systematically evaluate closed-source models and open-source models, with post-training analysis conducted on the latter. Our analysis reveals a visual-over-audio grounding asymmetry in open-source models and identifies distinct failure modes across SFT and RLVR. Together, these findings provide a model-specific diagnostic map to guide future research on omnimodal personalization.
\end{enumerate}

\begin{figure}[t]
    \vspace{-2em}

  \centering
  \includegraphics[width=\columnwidth]{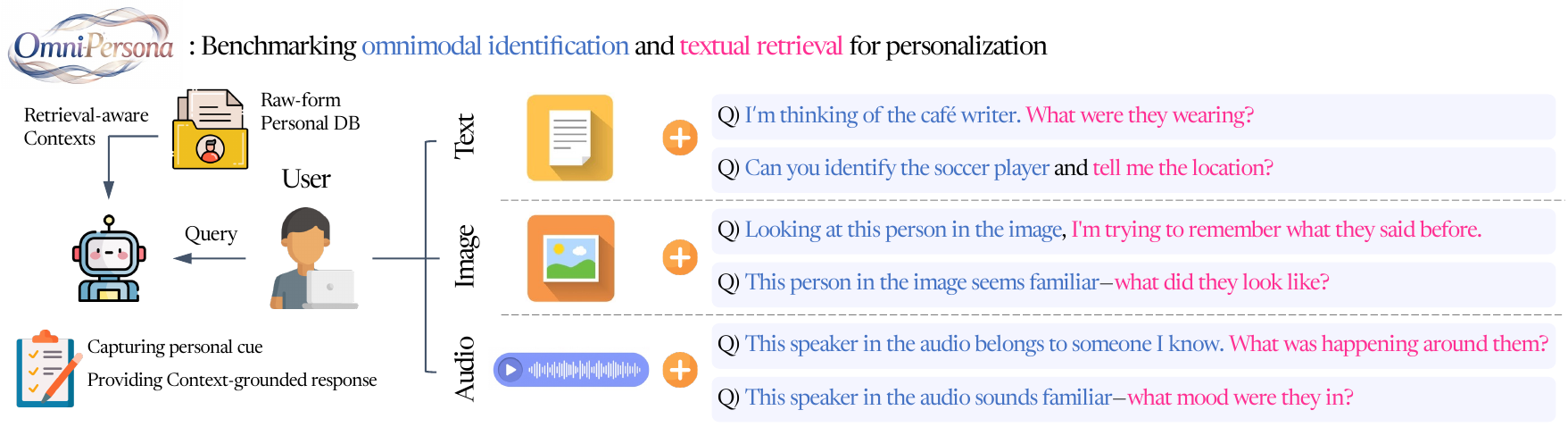}
    \caption{\textbf{Formulation of omnimodal personalization in the Omni-Persona benchmark.} As illustrated, a user query consists of a target modality and a text prompt, which we decompose into two main axes: \textcolor{omniblue}{blue} elements denote the process of perceptual identification, whereas \textcolor{myHotPink}{pink} elements highlight the realistic textual retrieval desired by the user. Note that our underlying assumption is that the raw-form contexts are pre-retrieved; we do not evaluate the retrieval process itself.}
\label{fig:omnimodal}
\end{figure}

\vspace{-1em}
\section{Related Works}
\textbf{Multimodal Personalization Methods.}
Early personalized vision-language models (VLMs)~\citep{myvlm, yollava, an2025unictokens} repurpose off-the-shelf models to recognize user-defined concepts via zero- or few-shot retrieval, yet remain brittle when new concepts must be incorporated dynamically into a user's memory. Post-training-based approaches subsequently emerged to mitigate this rigidity. \citet{rap} first demonstrated that SFT over retrieval-augmented user contexts enables coherent personalized response generation, but its reliance on costly large-scale caption annotations limits practical scalability. To alleviate this annotation burden, \citet{repic, covip} introduced RLVR-based methods, validating their utility in multi-concept image captioning~\citep{repic} and contextualized scenarios~\citep{covip}. Despite this progress, audio has received comparatively limited attention throughout this evolution: visual identity and biographical text~\citep{hong2025tameing, atmbench} have served as the predominant persona modalities, and speaker voice audio has rarely been integrated within a single unified personalization framework.

\textbf{Evaluation Protocols for Multimodal Personalization.}
The evaluation protocols accompanying these methods, including those of~\citep{yollava, repic, covip, myvlm}, rely heavily on recall-centric metrics. Such metrics primarily reward surface-level signals, such as name recall and contextual dialogue snippets, that can be directly reinforced during post-training. As a result, generation quality, calibration under absent-persona queries, and the trade-offs introduced by post-training remain largely unmeasured. This limitation is further compounded by existing benchmarks, which often operate under tightly controlled settings and abstract away realistic retrieval noise. To overcome such limitations, our benchmark unveils failure modes that are otherwise hidden beneath recall-only evaluation in multimodal personalization. Specifically, it exposes hallucination and over-abstention behaviors that conventional recall-centric metrics fail to capture. To the best of our knowledge, no prior work has unified interleaved omnimodal contexts, absent-persona evaluation, and a rigorous diagnostic protocol within a single benchmark. Further related work is in Appendix~\ref{appendix:related_works}.

\section{Problem Formulation}
\label{sec_problem}

As illustrated in Figure~\ref{fig:omnimodal}, we formally define \emph{omnimodal personalization}, extending the vision-language personalization paradigm~\citep{yollava, rap, repic, covip, myvlm, atmbench, hong2025tameing} to incorporate audio as a persona modality alongside vision and text.

\textbf{Formal Definition.} We formalize omnimodal personalization as follows. Let a user's personal memory be denoted by $\mathcal{M} = \{(v_i, a_i, t_i)\}_{i=1}^{N}$, where each entry is a triplet comprising a visual identity $v_i$, an audio sample $a_i$, and an associated text descriptor $t_i$. Specifically, $v_i$ may represent a profile image or an appearance snapshot, $a_i$ a 5--15s voice sample or a conversational recording, and $t_i$ dialogue or biographical information. Given a new query comprising a user prompt alongside visual, audio, and textual inputs $( v_q, a_q, t_q)$, the scenario determines which input provides the relevant cue. The relevant entries are retrieved from the memory $\mathcal{M}$ to construct the aggregated top-$K$ contexts $\left\{\mathcal{C}_{i}\right\}_{i=1}^{K}$ where $\mathcal{C}_i = (v_i, a_i, t_i)$. Following this retrieval, the model must, at inference time:
\begin{enumerate}[leftmargin=*]
\item \emph{Recognize} which specific entry $\mathcal{C}_j$ within the aggregated contexts $\left\{\mathcal{C}_{i}\right\}_{i=1}^{K}$ corresponds to the relevant query cue, while disregarding the irrelevant query modalities; and
\item \emph{Selectively extract and integrate} the specific details pertinent to the query from the associated text $t_j$ of the identified entry $\mathcal{C}_j$ into a contextually grounded response.
\end{enumerate}
The model must first accurately identify the relevant query cue and then ground its personalized response in the corresponding context. Furthermore, the retrieved contexts $\left\{\mathcal{C}_{i}\right\}_{i=1}^{K}$ arrive in an interleaved format, where the components of each entry $\mathcal{C}_i$ appear in an ordered sequence.

\textbf{Why Raw Context Matters.}
Previous textual-memory-based multimodal personalization works~\citep{atmbench, hong2025tameing, long2025seeing} convert multimodal signals into compact textual descriptions, creating an information bottleneck that can discard fine-grained identity cues that are difficult to verbalize, such as vocal timbre and facial geometry. We therefore derive personalization directly from \emph{raw omnimodal context}, preserving images and audio as perceptual signals rather than reducing them to text. Further details are provided in Appendix~\ref{app:extended_design}.

\textbf{Contextual Grounding over Retrieval.}
We decompose omnimodal personalization into two conceptually distinct sub-problems: (i) \emph{retrieval}, identifying which memories in a user's history are relevant to a given query, and (ii) \emph{contextual grounding}, integrating retrieved multimodal evidence into a faithfully personalized response. This work focuses on the latter, which our evaluation separates from retrieval quality by construction. We refer to the ability to identify query-relevant evidence and faithfully integrate the associated persona information into a response as \emph{grounding expressiveness}. Our goal is to define, measure, and systematically improve this capability. 

\textbf{Task Composition and Controlled Post-retrieval Evaluation.}
\label{app:post_retrieval_scope}
To evaluate grounding expressiveness under post-retrieval conditions, we provide all models with the same pre-retrieved, person-related contexts. The model must align the query with the relevant visual, textual, or acoustic evidence and ground its free-form response in the associated persona information when such evidence is available. The supplied contexts include visually similar identities and gender-matched voice samples as hard distractors, approximating retrieval noise while keeping the input fixed across models. Omni-Persona therefore evaluates post-retrieval grounding and generation, rather than standalone retrieval quality.

\textbf{Positioning within Retrieval-Augmented Personalization.}
Although Omni-Persona does not directly evaluate retrieval, its findings inform end-to-end retrieval-augmented personalization in two respects. First, they expose generator-side limitations that retrieval improvements alone cannot resolve, as correctly retrieved evidence remains ineffective if the model cannot interpret and use it. Second, they indicate whether the generator remains reliable under imperfect retrieval, using relevant evidence when available while avoiding unsupported personalization when it is absent.

\section{Benchmarking Omnimodal Perception and Retrieval}
\begin{wrapfigure}[9]{r}{0.36\textwidth} 
  \vspace{-\intextsep} 
  \centering
\includegraphics[width=0.36\textwidth]{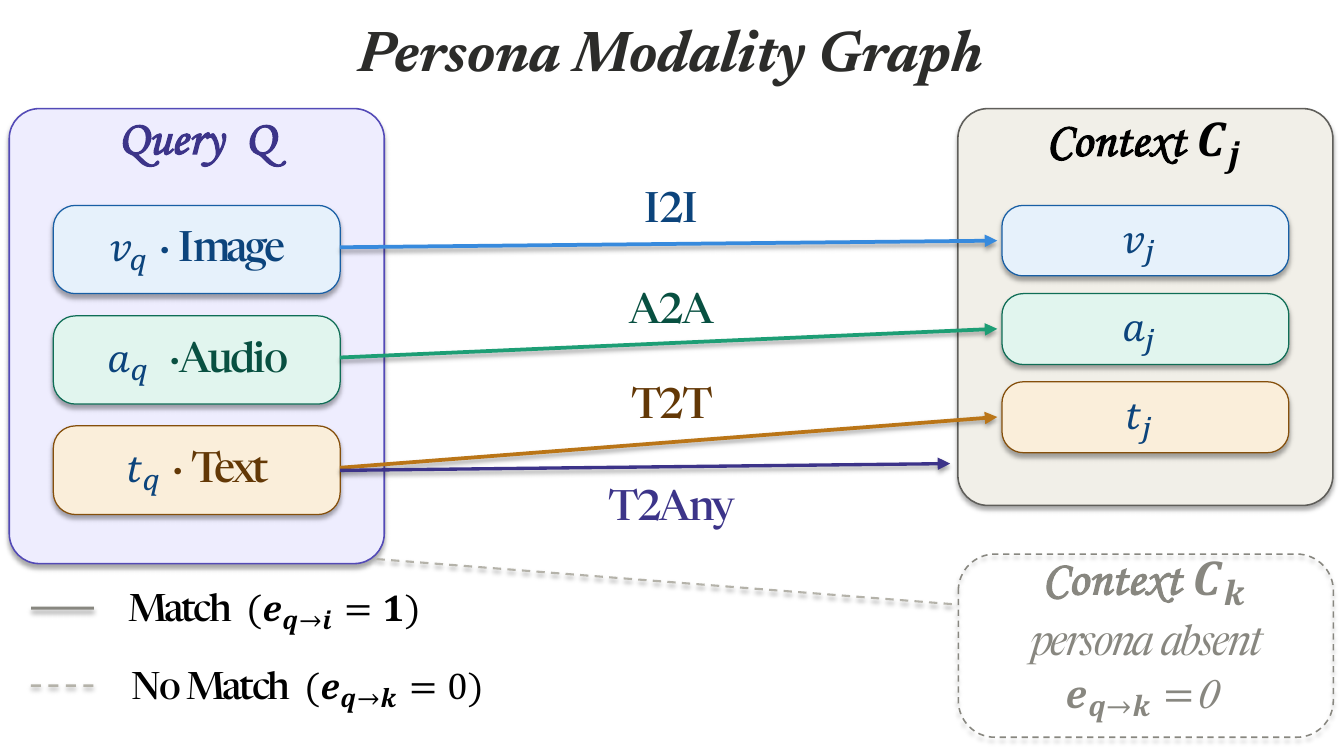}
  \vspace{-15pt} 
  \caption{PMG illustration.}
  \label{fig:pmg_plot}
  \vspace{-10pt} 
\end{wrapfigure}
We instantiate \textit{Omni-Persona} through the Persona Modality Graph (PMG), where each node is defined as a triplet representing an individual's omnimodal data. In this framework, personalization scenarios are modeled by the interconnections established between these nodes. Building upon this formulation, we propose a novel benchmark that simulates realistic personalization challenges, specifically focusing on modality matching (\textit{i.e.}, graph linkage) within the PMG.

\textbf{Persona Modality Graph (PMG) and Task Formulation.}
We formalize omnimodal personalization as a cross-modal routing problem over a PMG, $\mathcal{G} = (\mathcal{V}, \mathcal{E})$. The vertices $\mathcal{V}$ consist of a query node $\mathcal{Q}$ and retrieved context nodes $\mathcal{C}_1, \dots, \mathcal{C}_K$, where each node can encompass visual ($v$), audio ($a$), and textual ($t$) modalities, as represented in Figure~\ref{fig:pmg_plot}.

The core task is to determine whether a retrieved context contains the target persona and to establish a directed linkage (edge) $e_{q \to j} \in \mathcal{E}$ accordingly. Based on the provided query modality, we categorize the routing process into four primary matching scenarios:
\begin{enumerate}[label=(\arabic*), leftmargin=*]
    \item \emph{Image-to-Image (I2I)}: matching visual identity to an image query (\textit{i.e.}, visual identification);
    \item \emph{Audio-to-Audio (A2A)}: matching voice identity to an audio query (\textit{i.e.}, voice identification);
    \item \emph{Text-to-Text (T2T)}: matching textual attributes to a text query (\textit{i.e.}, same-modal semantic); and
    \item \emph{Text-to-Any (T2Any)}: aligning the semantic meaning of a text query with the cross-modal content of text, image, or audio (\textit{i.e.}, cross-modal semantic).
\end{enumerate}

Crucially, this formulation natively handles absent-persona calibration. If a context $\mathcal{C}_j$ contains the target persona, an active edge is formed ($e_{q \to j}=1$), making the grounded details in the associated text available to the model. Conversely, if the target persona is absent from all provided contexts, no edge is formed ($e_{q \to j}=0$), prompting the model to abstain confidently. This unified framework yields the four scenario groups in Table~\ref{tab:eval_scenario_summary_main} and 18 fine-grained tasks detailed in Appendix~\ref{appendix:data_construction}.

\begin{table}[t!]
\vspace{-1mm}
\centering
\footnotesize
\caption{\textbf{Omni-Persona evaluation scenario groups}. Each group tests whether a model can retrieve the correct persona information from a query in a different modality-matching setting.}
\label{tab:eval_scenario_summary_main}
\setlength{\tabcolsep}{6pt} 
\renewcommand{\arraystretch}{1.2}
\begin{tabularx}{\linewidth}{l c >{\hsize=1.4\hsize\raggedright\arraybackslash}X >{\hsize=0.6\hsize\raggedright\arraybackslash}X} 
\toprule
\textbf{Group} & \textbf{Setting} & \textbf{What the Model Must Match} & \textbf{Core Challenge} \\
\midrule
1: I2I & Image query & Match a person's face to the same visual identity in the persona context & Visual identity recognition \\
\addlinespace[2pt]
2: A2A & Audio query & Match a person's voice to the same speaker identity in the persona context & Voice identity recognition \\
\addlinespace[2pt]
3: T2T & Text query & Match a textual cue to the relevant textual persona attribute & Same-modal semantic retrieval \\
\addlinespace[2pt]
4: T2Any & Text query & Match the meaning of a text cue to relevant text, image, or audio information & Cross-modal semantic grounding \\
\bottomrule
\end{tabularx}
\vspace{-2mm}
\end{table}
\begin{figure}[t]
  \centering
\includegraphics[width=\columnwidth]{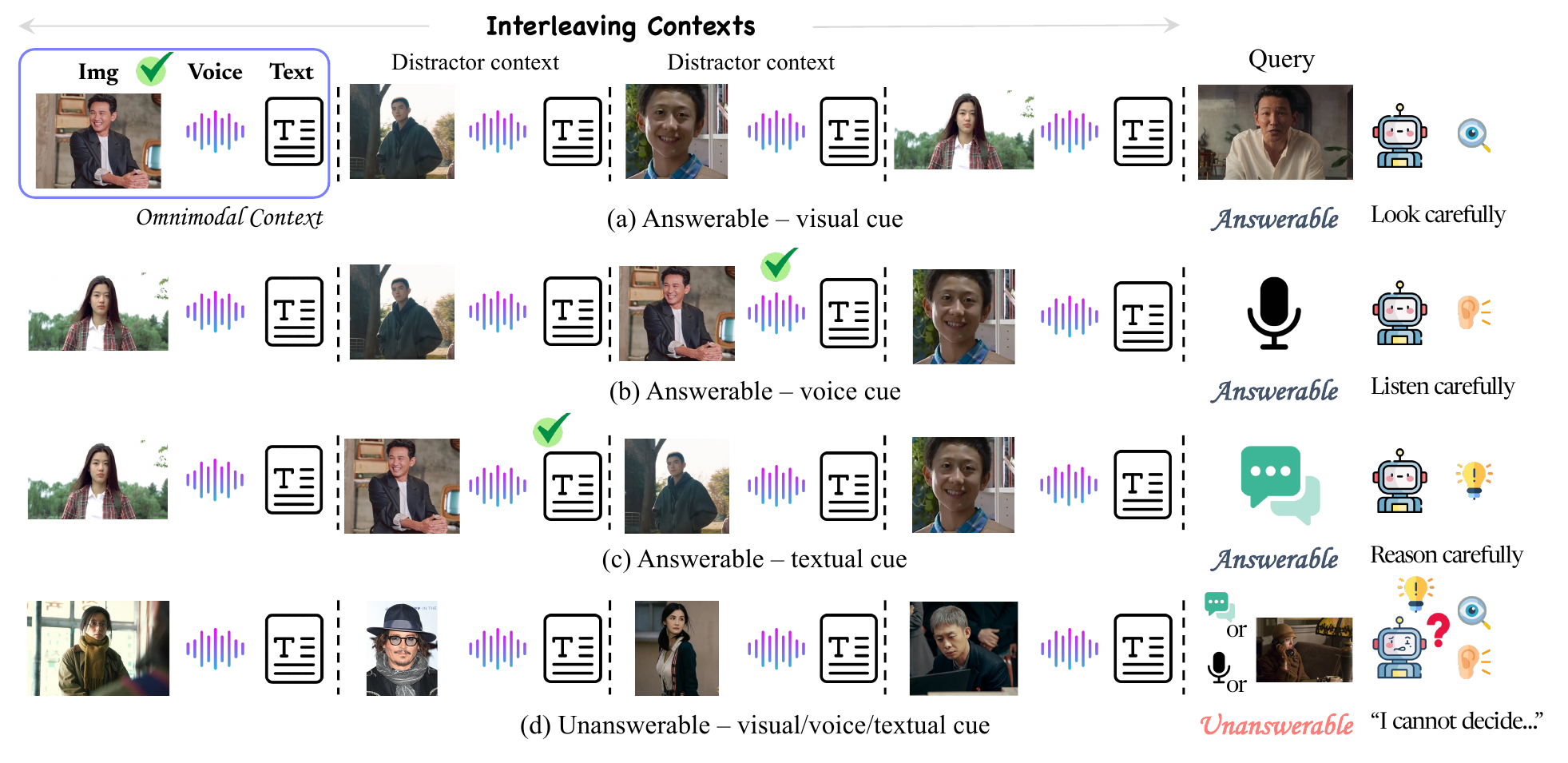}
\caption{Qualitative examples of context construction, personalization cues, distractors, and unanswerable cases in Omni-Persona.}
\label{fig:context}
\vspace{-2em}
\end{figure}

\textbf{Benchmark Design Principles.}
Designed around natural, human-centric interaction scenarios, Omni-Persona is, to our knowledge, the first personalization benchmark to incorporate \emph{audio} as a full persona modality alongside images and text, while systematically treating \emph{unanswerable} items---where the queried persona is absent from the retrieved context---as a primary evaluation dimension. Unlike prior personalization benchmarks~\citep{yollava, repic, covip}, which measure only recall---that is, whether the model retrieves the correct persona when present---Omni-Persona jointly evaluates \textit{grounding recall and abstention}. This setting reflects the dual challenge faced by real-world retrieval systems, where retrieval noise may cause the queried person to be absent from the retrieved context altogether. Furthermore, our cross-modal task design requires models to align audio evidence with visual descriptions, or vice versa, enabling the measurement of modality-specific grounding biases that unimodal tasks cannot reveal. Please see Appendix~\ref{app:extended_design} for further explanations.

\textbf{Robustness Under Retrieval Imperfection.}
Because real retrieval pipelines are noisy, Omni-Persona introduces two classes of perturbation into the evaluation benchmark. The first, \emph{hard distractors}, involves context entries from individuals who share visual or vocal similarities with the target. The second, \emph{no-GT retrieval}, omits the ground-truth persona from the context, demanding structured abstention instead of hallucinated matching. This rigorous setup guarantees a comprehensive evaluation across diverse omnimodal tasks. With approximately 50\% of the evaluation samples being no-GT, the benchmark systematically probes the model's resistance to hallucination, an essential desideratum when integrating with RAG systems~\citep{xu2025mem, chhikara2025mem0, li2026qwen3}.

\section{Experiments}
Our training study investigates which post-training regime most effectively aligns current omnimodal models for personalization. To this end, we systematically evaluate diverse models on our benchmark, elucidate the underlying behaviors surfaced by our evaluation metrics, and conduct an in-depth model debugging analysis to identify what is fundamentally required to advance omnimodal personalization. Details on data curation and implementation for the experiments are deferred to Appendix~\ref{appendix:training_details}.


\begin{table*}[t]

\definecolor{posgreen}{HTML}{00A36C} 
\definecolor{negred}{HTML}{FF0000}   

\vspace{-1.5mm}
\centering

\caption{
\textbf{Systematic Omni-Persona benchmark results.}
Complementary metrics include 1-FalseAbs and TrueAbs.
FA and TA stand for False Abstention and True Abstention, respectively.
\textcolor{posgreen}{Green} indicates improvement, and
\textcolor{negred}{Red} indicates regression relative to the base model after RLVR.
}
\label{tab:main_results}

\setlength{\aboverulesep}{0pt}
\setlength{\belowrulesep}{0pt}

\resizebox{\linewidth}{!}{%
\setlength{\tabcolsep}{4.5pt}
\renewcommand{\arraystretch}{1.2}

\begin{tabular}{@{}c l c c | ccc | ccc | ccc | ccc | ccc@{}}
\toprule

\multirow{2}{*}{}
  & \multirow{2}{*}{\textbf{Model}}
  & \multirow{2}{*}{\makecell{\textbf{Overall}\\\textbf{Ans}}}
  & \multirow{2}{*}{\makecell{\textbf{Overall}\\\textbf{Cal}}}
  & \multicolumn{3}{c|}{\textbf{I2I}}
  & \multicolumn{3}{c|}{\textbf{A2A}}
  & \multicolumn{3}{c|}{\textbf{T2T}}
  & \multicolumn{3}{c|}{\textbf{T2Any}}
  & \multicolumn{3}{c}{\textbf{Add. Metrics ($\times 100$)}} \\

\cmidrule(lr){5-7}
\cmidrule(lr){8-10}
\cmidrule(lr){11-13}
\cmidrule(lr){14-16}
\cmidrule(l){17-19}

  & & & &
  Ans & Unans & Cal &
  Ans & Unans & Cal &
  Ans & Unans & Cal &
  Ans & Unans & Cal &
  1-FA & TA & Avg \\

\midrule

\rowcolor{gray!15}
\cellcolor{white}
& \multicolumn{18}{l}{\textbf{Closed-source Models}} \\

& Gemini-3.6-Flash (Medium)
& \textbf{60.1} & \textbf{35.7}
& \textbf{71.3} & \textbf{6.0} & \textbf{38.5}
& \textbf{40.0} & \textbf{10.9} & \textbf{23.9}
& 69.9 & \textbf{9.8} & \textbf{50.0}
& 61.4 & \textbf{10.8} & 39.2
& 99.7 & \textbf{9.2} & \textbf{54.5} \\

& Gemini-3.6-Flash (High)
& 51.2 & 28.8
& 57.4 & 1.7 & 29.4
& 30.9 & 4.4 & 16.2
& \textbf{71.1} & 7.3 & \textbf{50.0}
& 49.4 & 7.7 & 31.1
& \textbf{100.0} & 4.5 & 52.2 \\

& Gemini-3.5-Flash-lite
& 51.4 & 29.7
& 65.2 & 5.2 & 35.1
& 12.7 & 8.0 & 10.1
& 62.7 & 2.4 & 42.7
& \textbf{72.3} & 6.2 & \textbf{43.2}
& 99.5 & 6.1 & 52.8 \\

\hline

\rowcolor{gray!15}
\cellcolor{white}
& \multicolumn{18}{l}{\textbf{Open-source Models}} \\

& MiniCPM-o 4.5 (Think)
& \textbf{41.7} & \textbf{39.3}
& 22.6 & \textbf{40.5} & \textbf{31.6}
& \textbf{17.3} & \textbf{38.7} & \textbf{29.1}
& \textbf{77.1} & 24.4 & \textbf{59.7}
& 65.1 & \textbf{33.8} & \textbf{51.4}
& 87.5 & \textbf{36.8} & \textbf{62.1} \\

& Phi-4 Multimodal
& 40.7 & 33.2
& \textbf{26.1} & 25.9 & 26.0
& 13.6 & 26.3 & 20.6
& 69.9 & 14.6 & 51.6
& \textbf{67.5} & 27.7 & 50.0
& \textbf{91.8} & 25.1 & 58.4 \\

& Interactive-Omni
& 32.7 & 29.6
& 21.7 & 19.8 & 20.8
& 9.1 & 31.4 & 21.5
& 55.4 & \textbf{29.3} & 46.8
& 56.6 & 24.6 & 42.6
& 84.9 & 26.2 & 55.5 \\

\midrule

& Qwen2.5-Omni-3B
& 34.0 & 36.7
& 27.8 & \textbf{44.0} & \textbf{35.9}
& 18.2 & \textbf{32.1} & \textbf{25.9}
& 45.8 & \textbf{53.7} & 48.4
& 51.8 & 38.5 & 45.9
& 74.9 & 39.6 & 57.2 \\

\rowcolor{cyan!8}
\cellcolor{white}
& \hspace{0.8em} + SFT (1K)
& 33.5 & 36.4
& 27.8 & \textbf{44.0} & \textbf{35.9}
& 17.3 & \textbf{32.1} & 25.5
& 44.6 & \textbf{53.7} & 47.6
& 51.8 & 38.5 & 45.9
& 74.9 & 39.6 & 57.2 \\

\rowcolor{cyan!8}
\cellcolor{white}
& \hspace{0.8em} + SFT (10K)
& 34.0 & \textbf{36.9}
& 26.1 & 43.1 & 34.6
& 18.2 & 31.4 & 25.5
& 45.8 & \textbf{53.7} & 48.4
& 54.2 & \textbf{44.6} & \textbf{50.0}
& 75.4 & \textbf{40.1} & \textbf{57.8} \\

\rowcolor{cyan!8}
\cellcolor{white}
& \hspace{0.8em} + RLVR
& \textbf{40.9} & 34.1
& \textbf{29.6} & 31.9 & 30.7
& \textbf{28.2} & 21.2 & 24.3
& \textbf{54.2} & 39.0 & \textbf{49.2}
& \textbf{60.2} & 21.5 & 43.2
& \textbf{84.7} & 26.7 & 55.7 \\

& \hspace{0.8em}\textcolor{black}{$\Delta$ vs.\ Base (RLVR)}
& \textcolor{posgreen}{+6.9}
& \textcolor{negred}{-2.5}
& \textcolor{posgreen}{+1.7}
& \textcolor{negred}{-12.1}
& \textcolor{negred}{-5.2}
& \textcolor{posgreen}{+10.0}
& \textcolor{negred}{-10.9}
& \textcolor{negred}{-1.6}
& \textcolor{posgreen}{+8.4}
& \textcolor{negred}{-14.6}
& \textcolor{posgreen}{+0.8}
& \textcolor{posgreen}{+8.4}
& \textcolor{negred}{-16.9}
& \textcolor{negred}{-2.7}
& \textcolor{posgreen}{+9.7}
& \textcolor{negred}{-12.8}
& \textcolor{negred}{-1.5} \\

\cmidrule(lr){2-19}

& Qwen2.5-Omni-7B
& 38.6 & 30.9
& 40.9 & \textbf{26.7} & 33.8
& 21.8 & \textbf{13.1} & 17.0
& 38.6 & \textbf{48.8} & 41.9
& 57.8 & \textbf{18.5} & 40.5
& 82.1 & \textbf{22.6} & 52.3 \\

\rowcolor{cyan!8}
\cellcolor{white}
& \hspace{0.8em} + SFT (1K)
& 39.6 & 31.5
& 40.9 & \textbf{26.7} & 33.8
& 21.8 & \textbf{13.1} & 17.0
& 42.2 & \textbf{48.8} & \textbf{44.4}
& 59.0 & \textbf{18.5} & 41.2
& 82.1 & \textbf{22.6} & 52.3 \\

\rowcolor{cyan!8}
\cellcolor{white}
& \hspace{0.8em} + SFT (10K)
& 42.7 & \textbf{31.9}
& \textbf{44.3} & 24.1 & \textbf{34.2}
& 22.7 & \textbf{13.1} & \textbf{17.4}
& 48.2 & 36.6 & \textbf{44.4}
& 61.4 & 16.9 & \textbf{41.9}
& 84.1 & 20.1 & 52.1 \\

\rowcolor{cyan!8}
\cellcolor{white}
& \hspace{0.8em} + RLVR
& \textbf{47.8} & 28.4
& 43.5 & 7.8 & 25.5
& \textbf{30.9} & 5.8 & 17.0
& \textbf{55.4} & 12.2 & 41.1
& \textbf{68.7} & 6.2 & 41.2
& \textbf{98.0} & 7.2 & \textbf{52.6} \\

& \hspace{0.8em}\textcolor{black}{$\Delta$ vs.\ Base (RLVR)}
& \textcolor{posgreen}{+9.2}
& \textcolor{negred}{-2.5}
& \textcolor{posgreen}{+2.6}
& \textcolor{negred}{-19.0}
& \textcolor{negred}{-8.2}
& \textcolor{posgreen}{+9.1}
& \textcolor{negred}{-7.3}
& \textcolor{posgreen}{+0.0}
& \textcolor{posgreen}{+16.9}
& \textcolor{negred}{-36.6}
& \textcolor{negred}{-0.8}
& \textcolor{posgreen}{+10.8}
& \textcolor{negred}{-12.3}
& \textcolor{posgreen}{+0.7}
& \textcolor{posgreen}{+15.9}
& \textcolor{negred}{-15.3}
& \textcolor{posgreen}{+0.3} \\

\cmidrule(lr){2-19}

\multirow{-11}{*}{\rotatebox{90}{\textbf{Qwen-Omni series}}}
& Qwen3-Omni-30B
& 51.2 & 33.7
& 53.0 & 21.6 & 37.2
& 25.5 & 8.0 & 15.8
& 65.1 & 19.5 & 50.0
& 68.7 & 13.8 & 44.6
& 94.4 & 14.8 & 54.6 \\

\midrule

& Gemma4-E2B
& 41.7 & \textbf{32.1}
& 45.2 & \textbf{7.8} & 26.4
& 15.5 & 43.8 & 31.2
& 56.6 & \textbf{4.9} & 39.5
& 56.6 & \textbf{10.8} & 36.5
& 88.7 & \textbf{21.7} & \textbf{55.2} \\

\rowcolor{cyan!8}
\cellcolor{white}
& \hspace{0.8em} + SFT (1K)
& 41.2 & 31.9
& 45.2 & \textbf{7.8} & 26.4
& 14.5 & 43.8 & 30.8
& 54.2 & \textbf{4.9} & 37.9
& 57.8 & \textbf{10.8} & \textbf{37.2}
& 88.7 & \textbf{21.7} & \textbf{55.2} \\

\rowcolor{cyan!8}
\cellcolor{white}
& \hspace{0.8em} + SFT (10K)
& 40.7 & 31.3
& 43.5 & 6.9 & 25.1
& 15.5 & \textbf{44.5} & \textbf{31.6}
& 53.0 & \textbf{4.9} & 37.1
& 57.8 & 7.7 & 35.8
& 89.0 & 21.2 & 55.1 \\

\rowcolor{cyan!12}
\cellcolor{white}
& \hspace{0.8em} + RLVR
& \textbf{44.5} & 28.1
& \textbf{48.7} & 6.0 & \textbf{27.3}
& \textbf{18.2} & 18.2 & 18.2
& \textbf{59.0} & 2.4 & \textbf{40.3}
& \textbf{59.0} & 6.2 & 35.8
& \textbf{96.7} & 10.3 & 53.5 \\

& \hspace{0.8em}\textcolor{black}{$\Delta$ vs.\ Base (RLVR)}
& \textcolor{posgreen}{+2.8}
& \textcolor{negred}{-4.0}
& \textcolor{posgreen}{+3.5}
& \textcolor{negred}{-1.7}
& \textcolor{posgreen}{+0.9}
& \textcolor{posgreen}{+2.7}
& \textcolor{negred}{-25.5}
& \textcolor{negred}{-13.0}
& \textcolor{posgreen}{+2.4}
& \textcolor{negred}{-2.4}
& \textcolor{posgreen}{+0.8}
& \textcolor{posgreen}{+2.4}
& \textcolor{negred}{-4.6}
& \textcolor{negred}{-0.7}
& \textcolor{posgreen}{+7.9}
& \textcolor{negred}{-11.4}
& \textcolor{negred}{-1.7} \\

\cmidrule(lr){2-19}

& Gemma4-E4B
& 46.8 & 31.2
& 50.4 & 19.0 & 34.6
& 15.5 & 17.5 & 16.6
& \textbf{62.7} & 0.0 & \textbf{41.9}
& 67.5 & \textbf{7.7} & \textbf{41.2}
& 92.8 & 14.2 & \textbf{53.5} \\

\rowcolor{cyan!8}
\cellcolor{white}
& \hspace{0.8em} + SFT (1K)
& 43.5 & 26.9
& 41.7 & 5.2 & 23.4
& 13.6 & 16.1 & 16.2
& 57.8 & \textbf{2.4} & 39.5
& 67.5 & 4.6 & 39.9
& \textbf{96.9} & 8.9 & 52.9 \\

\rowcolor{cyan!8}
\cellcolor{white}
& \hspace{0.8em} + SFT (10K)
& 46.8 & \textbf{33.7}
& \textbf{55.7} & \textbf{25.0} & \textbf{40.3}
& 13.6 & \textbf{26.3} & \textbf{20.6}
& 61.4 & \textbf{2.4} & \textbf{41.9}
& 63.9 & 6.2 & 38.5
& 85.7 & \textbf{19.5} & 52.6 \\

\rowcolor{cyan!12}
\cellcolor{white}
& \hspace{0.8em} + RLVR
& \textbf{48.6} & 30.4
& 52.2 & 14.7 & 33.3
& \textbf{18.0} & 13.1 & 15.4
& \textbf{62.7} & 0.0 & \textbf{41.9}
& \textbf{69.9} & 4.6 & \textbf{41.2}
& 94.4 & 10.6 & 52.5 \\

\multirow{-10}{*}{\rotatebox{90}{\textbf{Gemma4-series}}}
& \hspace{0.8em}\textcolor{black}{$\Delta$ vs.\ Base (RLVR)}
& \textcolor{posgreen}{+1.8}
& \textcolor{negred}{-0.8}
& \textcolor{posgreen}{+1.7}
& \textcolor{negred}{-4.3}
& \textcolor{negred}{-1.3}
& \textcolor{posgreen}{+2.5}
& \textcolor{negred}{-4.4}
& \textcolor{negred}{-1.2}
& \textcolor{gray}{+0.0}
& \textcolor{gray}{+0.0}
& \textcolor{gray}{+0.0}
& \textcolor{posgreen}{+2.4}
& \textcolor{negred}{-3.1}
& \textcolor{gray}{+0.0}
& \textcolor{posgreen}{+1.5}
& \textcolor{negred}{-3.6}
& \textcolor{negred}{-1.0} \\

\hline
\end{tabular}%
}

\vspace{-0.5em}
\end{table*}
\vspace{-2mm}
\subsection{Experimental Setup}
\textbf{Used Models.}
We evaluate four open-source omnimodal backbones (Gemma4-E2B/E4B-it and Qwen2.5-Omni-3B/7B) under three training regimes: SFT-1K, SFT-10K, and RLVR. We additionally include the closed-source Gemini-3 family~\citep{gemini3}, together with four open-source baselines: Qwen3-Omni-30B-A3B-Instruct~\citep{zhang2025qwen3}, Phi-4-multimodal-Instruct~\citep{phi4multimodal}, Interactive-Omni-8B~\citep{tong2025interactiveomni}, and MiniCPM-o 4.5~\citep{minicpmo}. 

\textbf{SFT Training Setup.}
We construct a $10\text{K}$-sample SFT dataset spanning 12 distinct task types, complemented by a $1\text{K}$ subset for efficient ablation studies. This corpus encompasses foundational grounding, audio-centric scenarios, and absent-persona cases designed to promote calibrated abstention. Crucially, we curate this dataset for broad modality alignment across image, audio, and text, rather than narrow, benchmark-specific optimization. We emphasize that constructing a training corpus for SFT is fundamentally constrained by several factors: (i) the \textit{inherent noise} in synthesizing high-quality ground truth responses for diverse personalization scenarios~\citep{rap, repic}; (ii) the \textit{unpredictability} of the test-time query distribution~\citep{repic, jiang2025know, hong2025tameing}; and (iii) the \textit{scarcity} of large-scale, paired real-world multimodal data~\citep{hu2025investigating, liu2025nexus}, necessitating a reliance on synthetic samples that may introduce domain bias. These limitations collectively make SFT alone insufficient for ensuring predictable personalization coverage at test time, consistent with recent omnimodal post-training studies showing that RL-based objectives can substantially outperform SFT under matched data and compute budgets~\citep{rouditchenko2025omni, li2025reinforcement, yang2025humanomniv2}. Consequently, we treat SFT as a comparative baseline for RLVR to rigorously analyze how the two post-training regimes shape performance trends.

\textbf{RLVR Training Setup.}
\label{sec:targeted_rl}
To ensure a fair comparison, we perform RLVR on synthetic persona contexts produced by the same pipeline as the SFT corpus. Thus, both regimes use the same type of image--audio--text context triplets and queries. Unlike SFT, however, RLVR does not require reference GT responses. Instead, the model is trained with verifiable feedback that checks whether its response satisfies the intended capability, such as perceptual matching or grounded retrieval. We use three complementary reward components:

\vspace{-0.5em}
\begin{enumerate}[label=(\arabic*), leftmargin=*]
    \item \textit{Perception (Rule-based):} Evaluates visual or auditory persona matching by comparing the model's binary decision (\textit{i.e.}, \texttt{yes}/\texttt{no}) with the GT label derived from the query--context pairing.

    \item \textit{Location (Rule-based):} Evaluates whether the model correctly identifies the GT context (\textit{i.e.}, MCQA) index, thereby encouraging precise localization within the context.
    
    \item \textit{Retrieval (LLM-as-a-Judge):} Evaluates whether the response is grounded in retrieved persona evidence. When the target persona is present, the judge verifies factual support against the GT answer and context; when absent, it checks whether the model correctly abstains.
\end{enumerate}
\vspace{-1em}

For the retrieval reward, we use \texttt{GPT-5.4}~\citep{singh2025openai} to generate training queries and GT answers from the benchmark scenarios, rather than full reference responses, while enforcing a disjoint split to prevent evaluation overlap (see Figure~\ref{fig:rlvr_overview} for an example of the RLVR training framework). 

\begin{figure}[t]
  \centering
  \vspace{-1em}
\includegraphics[width=\columnwidth]{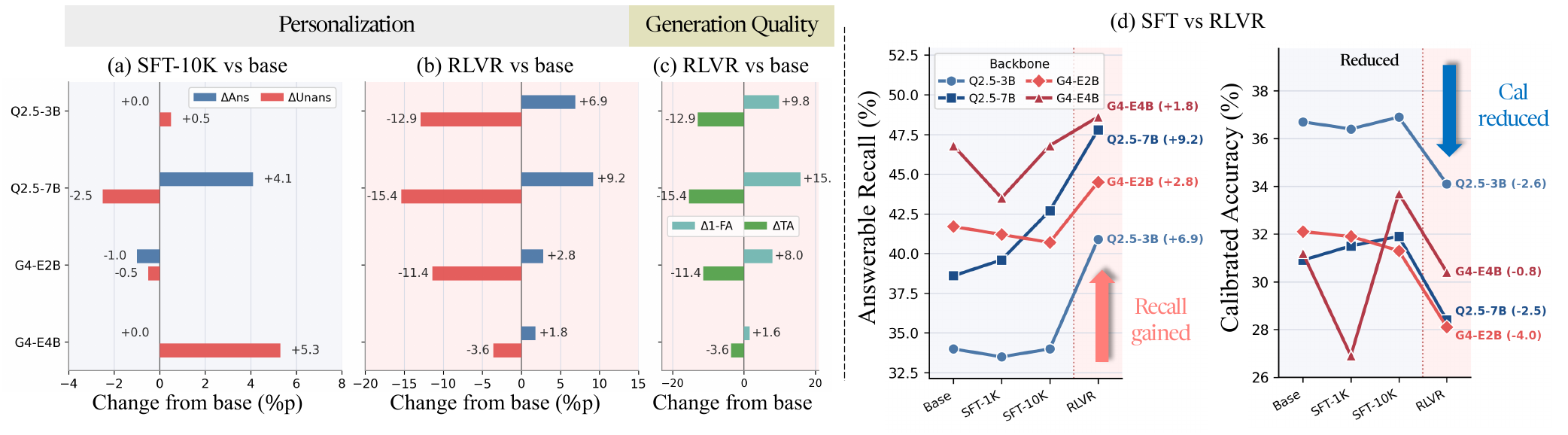}
\caption{\textbf{Performance across post-training regimes}. (a) SFT scaling does not directly translate to gains in open-ended personalization scenarios.
(b) RLVR directly improves answerable recall over SFT.
(c) RLVR shifts models toward answering, broadly decreasing abstention ($1-\mathrm{FA}$ up, $\mathrm{TA}$ down).
(d) RLVR exhibits a recall--calibration trade-off, lowering $\mathrm{Cal}$ below the base model.}
\label{fig:sft_rl_compare}
\vspace{-1em}
\end{figure}

\textbf{Evaluation Metrics.}
We treat visual and auditory perception jointly: a model that \textcolor{black}{\emph{``sees, hears, and reasons well''}} must \textcolor{black}{\emph{consistently match identical entities and distinguish distinct ones across both modalities.}} To rigorously evaluate these capabilities, a faithful assessment of omnimodal personalization must simultaneously account for two critical failure modes that standard recall metrics conflate: \emph{hallucinating an identity absent from the context} and \emph{wrongly answering when the identity is absent}. Accordingly, we evaluate models on a fixed pool of 750 queries, balanced between answerable and unanswerable scenarios. Further judge-reliability analyses are detailed in Appendix~\ref{appendix:judge_reliability}. Our evaluation relies on the following key metric groups:
\begin{itemize}[leftmargin=*]
    \item \textit{Calibrated Accuracy}: Our primary metric is the item-weighted accuracy over all queries, $\mathrm{Cal} = \tfrac{n_{\mathrm{a}}\mathrm{Ans} + n_{\mathrm{u}}\mathrm{Unans}}{n_{\mathrm{a}} + n_{\mathrm{u}}}$, where $n_{\mathrm{a}}$ and $n_{\mathrm{u}}$ are the numbers of answerable and unanswerable items. \emph{Answerable Recall} ($\mathrm{Ans}$) is evaluated for binary correctness using an LLM-as-a-judge \texttt{GPT-5.4-mini}, while \emph{Unanswerable Recall} ($\mathrm{Unans}$) is measured via abstention-keyword matching~\citep{chen2024benchmarking}. $\mathrm{Cal}$ serves to operationalize a model's true expressiveness and grounding in omnimodal personalization.
    
    \item\textit{Anti-Hallucination Accuracy:} To prevent models from artificially inflating $\mathrm{Cal}$ through blanket abstention, we rigorously assess their anti-hallucination reliability. Specifically, we report two complementary metrics: the answerable-side complement of False Abstention ($1-\mathrm{FA}$), and the True Abstention ($\mathrm{TA}$) rate on unanswerable items. These metrics capture generation quality, complementing $\mathrm{Cal}$, and higher values indicate better performance across all reported metrics.
\end{itemize}

\subsection{Main Results}
Table~\ref{tab:main_results} reports $\mathrm{Cal}$ scores across all models and benchmark scenarios. Building on these results, Figure~\ref{fig:sft_rl_compare} compares post-training regimes in terms of recall, calibration, and abstention behavior.

\textbf{Closed-Source Models.}
The $\mathrm{Cal}$ score exposes a sharp divide that recall alone hides. No closed-source model achieves both strong grounding and reliable abstention. Gemini-3.6-Flash leads in $\mathrm{Ans}$ (60.1\%) but collapses on Unans, yielding only 35.7\% $\mathrm{Cal}$. This low $\mathrm{Cal}$ mainly reflects hallucination on absent-persona items, evidenced by its especially low TA score.

\textbf{Open-Source Baselines.}
Among open-source models, parameter scale alone does not guarantee high $\mathrm{Cal}$ in our benchmark: Q3-30B \emph{underperforms} Q2.5-3B on $\mathrm{Cal}$ (33.7\% vs.\ 36.7\%), and Phi-4 trails the 3B Qwen baseline, while MiniCPM attains the highest open-source $\mathrm{Cal}$ (39.3\%) via strong abstention. Among our four RLVR backbones, Q2.5-3B retains the highest $\mathrm{Cal}$ score (34.1\%). We further discuss the discrepancy between scaling and $\mathrm{Cal}$ score in Section~\ref{sec:analysis} (Key Finding 2).

\textbf{Scaling SFT Fails to Bridge the Distribution Gap in Open-Ended Personalization.}
Expanding the SFT dataset from $1\text{K}$ to $10\text{K}$ samples fails to yield consistent $\mathrm{Cal}$ improvements and even leads to performance degradation (Figure~\ref{fig:sft_rl_compare}a,d) for the Gemma4 series. While the augmented corpus was curated to strengthen omnimodal matching across diverse contexts, this broader coverage does not translate into superior benchmark performance. This discrepancy highlights the rigorous complexity of the open-ended personalization required at test time and provides empirical evidence for the inherent limitations of SFT data construction: simply scaling training volume for general modality alignment is insufficient to bridge the distributional gap in complex, personalized reasoning tasks.

\textbf{RLVR Delivers Recall Gains through Verifiable Supervision.}
Conversely, RLVR directly and substantially improves answerable recall (Figure~\ref{fig:sft_rl_compare}b), extending the findings of RePIC and CoViP~\citep{repic, covip} to the omnimodal personalization domain. Recall improves on every backbone, with the largest gains on the Qwen models (Ans $+6.9$ for 3B and $+9.2$ for 7B) and smaller gains on Gemma ($+2.8$ and $+1.8$). However, these gains do not translate into calibration: RLVR fails to induce abstention, and $\mathrm{Cal}$ falls below base for every backbone ($-$0.8 to $-$4.0); MiniCPM-o 4.5 retains the highest $\mathrm{Cal}$ and anti-hallucination average ($1-\mathrm{FA}$ and $\mathrm{TA}$) among the evaluated models, as shown in Table~\ref{tab:main_results}. The detailed analyses of the recall--$\mathrm{Cal}$ trade-off are in Appendix~\ref{sec:app_additional_analysis}, including paired significance tests and confidence intervals for this trade-off in Appendix~\ref{sec:app_significance}.

\textbf{The Answer-Inducing Trade-off of RLVR.}
Notably, our anti-hallucination accuracies reveal that \textit{RLVR shifts the backbones toward answering}. As shown in Figure~\ref{fig:sft_rl_compare} (c), all four backbones raise $1-\mathrm{FA}$ and drop $\mathrm{TA}$ below the baseline (most pronounced for the Qwen backbones), indicating fewer false abstentions on answerable cases but a collapse of abstention on unanswerable ones. We interpret this as a \emph{reward-induced answering bias}: since our outcome rewards directly target grounding and reasoning, attempting a grounded answer becomes the dominant signal even when the target persona is absent, so abstention is rarely reinforced. This effect appears across model families and underlies the $\mathrm{Cal}$ decline observed across the backbones. More sophisticated reward shaping or cold-start SFT~\citep{guo2025deepseek} may restore calibrated abstention, which we leave to future work.

\vspace{-1em}

\section{Toward Omnimodal Personalization: In-Depth Analyses}
\label{sec:analysis}

\begin{keyfinding}{Key Finding 1: Audio Grounding Lags Behind Visual Grounding}
Open-source models consistently underperform on audio grounding relative to visual grounding, exhibiting an approximately 5--35 percentage-point gap in answerable recall between A2A and I2I tasks. Although RLVR improves recall in the audio modality, $\mathrm{Cal}$ reveals a key limitation: models trained with naive RLVR are biased away from abstention. 
\end{keyfinding}

When evaluated separately on visual and auditory tasks, open-source models exhibit substantially stronger visual than auditory perception, resulting in a pronounced performance gap between I2I and A2A tasks. Although the closed-source Gemini models achieve the strongest A2A performance, reaching up to 40.0\%, they still exhibit a clear cross-modal perceptual gap. By providing explicit perceptual supervision through rule-based rewards, RLVR improves A2A answerable recall across all four backbones. The gains are substantial for the Qwen models, reaching $+10.0$ and $+9.1$ points for the 3B and 7B variants, respectively, while G4-E2B and G4-E4B improve by $+2.7$ and $+2.5$ points, respectively. Notably, the A2A recall gains for Qwen exceed those observed on I2I, demonstrating that RLVR can effectively elicit audio-perception capabilities. However, $\mathrm{Cal}$ reveals a key limitation of RLVR alone: without an SFT warm-up, the models become biased against abstention, leaving the trade-off between recall and calibrated abstention only partially resolved. Consequently, RLVR yields lower $\mathrm{Cal}$ scores than the base models on I2I and A2A tasks for most backbones.

\begin{keyfinding}{Key Finding 2: Recall and Model Scale Miss the Calibration Axis}
Answerable recall and model scale are useful but incomplete diagnostics for personalization. A model with strong recall may still hallucinate when the target persona is absent (low $\mathrm{TA}$), or over-abstain even when the relevant persona is present (low $1-\mathrm{FA}$). Likewise, $\mathrm{Cal}$ exposes this missing axis by jointly measuring grounding on answerable cases and abstention on unanswerable cases.
\end{keyfinding}

A closer look at Q3-30B illustrates this point. While the 30B model achieves higher answerable recall on text-grounded tasks, it performs worse on unanswerable cases than smaller Q2.5 variants, resulting in lower overall $\mathrm{Cal}$ and the lowest $\mathrm{TA}$ score among the Qwen models, indicating greater hallucination. Similarly, Gemini-3.6-Flash achieves the strongest answerable recall (Ans 60.1\%) yet pairs it with a low $\mathrm{TA}$ (9.2\%), capping its calibrated accuracy at 35.7\%. Increasing its thinking budget further degrades performance, suggesting that additional reasoning does not necessarily improve personalization. Recall-only comparisons often obscure these discrepancies, whereas $\mathrm{Cal}$ and anti-hallucination accuracy effectively expose them.

\begin{keyfinding}{Key Finding 3: SFT and RLVR Reveal Different Post-training Trade-offs}
SFT and RLVR fail in different ways. SFT is limited by the difficulty of constructing annotated GT supervision at scale, making it challenging to consistently improve evaluation performance. In contrast, RLVR can directly target perceptual grounding and textual retrieval through verifiable reward signals, yielding answerable-recall gains on every backbone, largest for the Qwen models. However, it fails to induce calibrated abstention: the backbones shift toward answering ($1-\mathrm{FA}$ up, $\mathrm{TA}$ down), and $\mathrm{Cal}$ fails to improve for any backbone.
\end{keyfinding}

SFT is limited by the difficulty of constructing high-quality supervision data at scale. Scaling the data from 1K to 10K does not reliably improve $\mathrm{Cal}$, suggesting that a broader data mixture does not directly translate into better benchmark performance in open-ended personalization scenarios. This can be interpreted as a training-evaluation mismatch~\citep{chu2025sft}. Conversely, RLVR avoids this bottleneck by replacing reference-response imitation with outcome-level supervision. Perception reward reinforces binary visual/audio matching, location reward reinforces correct localization within a context, while retrieval reward reinforces responses grounded in textual evidence. This enables RLVR to substantially improve answerable recall, especially for the Qwen backbones, while bypassing the need for GT responses. However, because these rewards directly target grounding and reasoning, the models drift toward answering (higher $1-\mathrm{FA}$, lower $\mathrm{TA}$), and calibrated abstention is not induced: $\mathrm{Cal}$ fails to improve for any backbone. Overall, these analyses reaffirm the potential of RLVR-based frameworks while highlighting the need for careful post-training to mitigate the inherent trade-off between recall and calibrated abstention.
 

\vspace{-0.2em}
\section{Conclusion}
We introduce \emph{Omni-Persona}, the first comprehensive benchmark for omnimodal personalization. By formalizing contextual grounding over retrieved persona evidence and the integration of raw-form omnimodal context, it enables systematic analysis of personalized expressiveness, treats audio as a persona modality alongside images and text, and adopts absent-persona as a core evaluation dimension. We propose calibrated accuracy and anti-hallucination accuracy, showing that recall-only metrics can obscure hallucination under retrieval noise. Extensive benchmarking shows that, on omnimodal tasks evaluated independently, open-source models process visual cues substantially more reliably than audio cues; scaling SFT does not reliably improve performance, reflecting the difficulty of constructing data aligned with open-ended personalization; and RLVR substantially improves recall but fails to induce calibrated abstention, leaving calibrated accuracy below base. Overall, Omni-Persona provides a realistic diagnostic framework for analyzing the strengths and failure modes of omnimodal personalization.

\textbf{Limitations.} Our benchmark uses synthetic audio and text with rigorous model-based filtering, leaving further human verification as a future refinement. Free-form LLM-as-a-judge evaluation may introduce residual bias; see Appendix~\ref{appen:limitations} for further discussion.



\section{Acknowledgements}
This work was supported by the NVIDIA Academic Grant Program; the Institute of Information \& Communications Technology Planning \& Evaluation (IITP) grant [No. RS-2021-II211343, Artificial Intelligence Graduate School Program (Seoul National University)] and the National Research Foundation of Korea (NRF) grant (No. 2022R1A3B1077720), both funded by the Korea government (MSIT); and the BK21 FOUR program of the Education and Research Program for Future ICT Pioneers, Seoul National University in 2026. This paper was also the result of the research project supported by SK hynix Inc.



\bibliographystyle{iclr2027_conference}
\bibliography{references}

\newpage
\appendix
\renewcommand{\thefigure}{S.\arabic{figure}}
\renewcommand{\thetable}{S.\arabic{table}}
\renewcommand{\theequation}{S.\arabic{equation}}

\setcounter{figure}{0}
\setcounter{table}{0}  
\setcounter{equation}{0}  

\onecolumn

\begin{table}[t!]
\centering
\caption{Comparison of previous personalization benchmarks against real-world scenarios.}
\label{tab:synthetic_vs_real}
\footnotesize
\setlength{\tabcolsep}{4pt}
\renewcommand{\arraystretch}{1.3} 

\begin{tabularx}{\linewidth}{@{} 
    >{\raggedright\arraybackslash}p{1.9cm} 
    >{\hsize=1.3\hsize\raggedright\arraybackslash}X 
    >{\hsize=0.9\hsize\raggedright\arraybackslash}X 
    >{\hsize=0.9\hsize\raggedright\arraybackslash}X 
    c
    c 
    c 
    >{\hsize=0.9\hsize\raggedright\arraybackslash}X 
    @{}}
\toprule
\textbf{Dataset} &
\textbf{Context Nature} &
\textbf{Acqui\-si\-tion} &
\textbf{Sce\-nar\-ios} &
\textbf{Tasks} &
\textbf{Modalities} &
\textbf{Cost} &
\textbf{Eval\-u\-a\-tion} \\
\midrule

CoViP~\citep{covip}
& Ho\-mo\-ge\-neous
& Synthetic only
& Text dialogue
& 3
& I, T
& Low (API)
& Standard (MCQA) \\

Omni-Persona \newline (Ours)
& Het\-ero\-ge\-neous \newline (distractors, no GT)
& Mixed \newline (Real \& Syn.)
& Dialogue, \newline bi\-og\-ra\-phy
& 18
& I, T, A
& Moderate
& Multi-hop \newline (Free form) \\

Real-world
& Noisy \& \newline in\-com\-plete
& Manual \newline (Privacy risks)
& Emails, logs, \newline his\-to\-ries, \newline memory update
& $\gg 18$
& I, T, A, V
& Prohibitive
& Complex \newline (Agentic) \\

\bottomrule
\end{tabularx}
\end{table}
\vspace{-1em}

\section{Further Related Works}\label{appendix:related_works}
\vspace{-0.5em}
\textbf{LLM Personalization.}
Recent text-only personalization benchmarks evaluate how well LLMs profile and respond to users. For instance, \citet{jiang2025know} focus on dynamic profiling, \citet{kim2025cupid} use real user interaction logs for alignment, and \citet{jiang2025personamemv2} infer latent traits from conversational history. While foundational, these works remain confined to the textual modality, lacking the detailed analyses for multimodal grounding necessary for omnimodal personal assistants.

\textbf{Comparison with CoViP.}
Table~\ref{tab:synthetic_vs_real} contrasts Omni-Persona with CoViP~\citep{covip}, its closest predecessor in the multimodal personalization benchmark space. Relative to CoViP, Omni-Persona: (i) incorporates audio as a persona modality alongside image and text; (ii) incorporates ground-truth-absent queries under multi-distractor contexts as a core evaluation axis; and (iii) broadens the task scope from image captioning to open-ended QA and cross-modal identity matching. Critically, CoViP's answerable-only multi-choice question answering (MCQA) protocol cannot surface model overconfidence or abstention collapse. By exposing these failure modes through a dual-axis design and extending absent-persona evaluation to audio and cross-modal scenarios, Omni-Persona offers a comprehensive diagnostic framework for omnimodal personalization.


\section{Preliminaries}
\textbf{Omnimodal and Multimodal Foundation Models.}
Omnimodal models extend MLLMs toward any-to-any generation across text, vision, and audio. Proprietary systems such as GPT-4o~\citep{openai2024} and Gemini-3~\citep{gemini3} have shown strong cross-modal understanding and real-time dialogue capabilities. Among open-source omnimodal models, Qwen2.5-Omni~\citep{qwen25omni} adopts a Thinker--Talker architecture: the Thinker performs multimodal understanding and textual reasoning, while the Talker generates speech responses. MiniCPM-o~\citep{minicpmo} unifies vision, speech, and language within an end-to-end framework, employing a time-division multiplexing mechanism that interleaves parallel omnimodal streams into periodic time slices, thereby supporting real-time bidirectional speech interaction. Phi-4-Multimodal~\citep{phi4multimodal} integrates text, vision, and speech/audio into a single 5.6B-parameter model through a Mixture-of-LoRAs design, in which modality-specific LoRA adapters and routers are attached to a frozen Phi-4-Mini backbone, enabling competitive multimodal reasoning without cross-modal interference while fully preserving the base language capability. Interactive-Omni~\citep{tong2025interactiveomni} combines omni-modal pre-training with interaction-oriented post-training to support text-and-speech generation in multi-turn audio-visual dialogue with long-term contextual memory.

We further use Gemma4 E-series models (E2B and E4B)~\citep{gemma4} as our main analysis models. Gemma4 is not a native omnimodal model, but an MLLM designed for constrained DRAM and flash-memory environments through Per-Layer Embeddings and Grouped-Query Attention, which reduce KV-cache pressure. This makes it a useful on-device-oriented counterpart to larger server-class omnimodal systems.


\section{Benchmark Design Rationale and Integrity}
\label{app:extended_design}
\textbf{Construction and Data Provenance.}
\label{app:data_provenance}
The benchmark comprises approximately 750 instances spanning 18 fine-grained tasks. Training images are synthetic, while training and evaluation audio are drawn from disjoint source splits to prevent contamination. Each instance is manually verified after constructing person-consistent query--context histories and introducing modality-appropriate hard distractors. To prevent text-only shortcuts in the visual- and voice-identity tasks, we ensure that all four persona descriptions state the queried subfield, each with a distinct, gender-consistent value. We further use \texttt{GPT-5.5} to rewrite the descriptions in a uniform natural style while preserving the gold semantics. This construction removes cues based on attribute presence, template structure, or writing style, requiring the model to identify the relevant persona from the query image or audio.

\textbf{Role of Absent-persona Evaluation.}
\label{app:absent_persona_role}
Approximately half of the instances omit the target persona from the context, modeling retrieval failures in which relevant evidence is unavailable. These cases test whether the model abstains rather than producing an unsupported answer from a distractor, with failure patterns evaluated across visual, acoustic, and textual evidence. Answerable recall alone does not capture this behavior or distinguish appropriate abstention from systematic over-abstention. We therefore jointly report Cal, $1-\mathrm{FA}$, and TA to evaluate grounding and abstention behavior.

\textbf{Mitigating Textual Shortcuts in Contextual Memory.}
\label{app:textual_shortcut}
We further enforce within-description consistency in names, gender references, and pronouns. Topic- and quote-related candidates are also expressed in the same neutral third-person register, removing first-person or shared-memory cues that could distinguish the gold context. These controls minimize residual lexical and stylistic asymmetries between the gold persona and distractors.

\begin{table*}[h!]
    \centering
    \small
    \renewcommand{\arraystretch}{1.0}
    \setlength{\tabcolsep}{6pt}
    \caption{Modality ablation measured by strict answerable recall across task groups. Parentheses in the \emph{Both dropped} rows indicate changes relative to the corresponding full-context results.}
    \label{tab:app_modality_ablation}
    \begin{tabular}{lrrrrr}
        \toprule
        Condition & I2I & A2A & T2T & T2Any & $1-\mathrm{FA}$ \\
        \midrule

        \multicolumn{6}{l}{\textbf{Qwen2.5-Omni-3B}} \\
        Full context
            & 28 & 19 & 45 & 52 & 75 \\
        Image shuffled
            & 12 & 13 & 33 & 49 & 68 \\
        Audio shuffled
            & 22 & 11 & 41 & 57 & 80 \\
        Image dropped
            & 8 & 14 & 27 & 41 & 44 \\
        Audio dropped
            & 50 & 11 & 52 & 65 & 80 \\
        \rowcolor{blue!7}
        \textbf{Both dropped}
            & \textbf{3 ($-25$)}
            & \textbf{4 ($-15$)}
            & \textbf{50 ($+5$)}
            & \textbf{62 ($+10$)}
            & \textbf{26 ($-49$)} \\

        \addlinespace[3pt]
        \midrule
        \multicolumn{6}{l}{\textbf{Gemma4-E2B}} \\
        Full context
            & 44 & 15 & 57 & 57 & 89 \\
        Image shuffled
            & 12 & 12 & 49 & 49 & 81 \\
        Audio shuffled
            & 39 & 9 & 54 & 63 & 92 \\
        Image dropped
            & 4 & 14 & 48 & 46 & 72 \\
        Audio dropped
            & 57 & 10 & 63 & 64 & 89 \\
        \rowcolor{blue!7}
        \textbf{Both dropped}
            & \textbf{3 ($-41$)}
            & \textbf{2 ($-13$)}
            & \textbf{66 ($+9$)}
            & \textbf{72 ($+15$)}
            & \textbf{47 ($-42$)} \\

        \bottomrule
    \end{tabular}
\end{table*}

\textbf{Modality Corruption and Grounding-Dependence Ablation.}
\label{app:modality_ablation}
In Table~\ref{tab:app_modality_ablation}, we empirically test whether performance remains dependent on visual and auditory evidence after controlling for textual shortcuts. The textual inputs are held fixed, while the selected perceptual modality is either shuffled or removed. Under \emph{shuffling}, the query and context inputs are replaced with those associated with different personas; under \emph{dropping}, the selected modality is removed entirely. We evaluate Qwen2.5-Omni-3B and Gemma4-E2B, reporting strict answerable recall by task group together with $1-\mathrm{FA}$.

The perceptual tasks show clear modality-specific dependence. Removing both image and audio reduces I2I recall from 28 to 3 for Qwen2.5-Omni-3B and from 44 to 3 for Gemma4-E2B; A2A recall similarly decreases from 19 to 4 and from 15 to 2, respectively. The targeted ablations support the same conclusion: image corruption primarily degrades I2I, whereas audio corruption degrades A2A. Because the textual inputs remain unchanged, these losses show that the perceptual tasks cannot be recovered from text alone.

Removing both modalities also reduces $1-\mathrm{FA}$ from 75 to 26 for Qwen2.5-Omni-3B and from 89 to 47 for Gemma4-E2B, indicating increased false abstention when perceptual evidence is unavailable. In contrast, T2T and T2Any improve for both models after removing perceptual inputs, suggesting that irrelevant image and audio inputs can interfere with text-grounded retrieval tasks. Overall, the ablation confirms that \textit{performance depends on the modality relevant to each task rather than on a shared textual shortcut}.

\section{Details on Evaluation Dataset and Metrics}\label{appen:metrix}
\paragraph{Task Overview.} Omni-Persona instantiates the four PMG scenarios into 18 fine-grained sub-tasks organized into four distinct groups, each covering a different combination of query and context modalities. The groups are structured as follows:

\begin{itemize}[leftmargin=*, itemsep=4pt]
    \item \emph{Group 1 (I2I; 5 tasks):} A face-image query is matched against the persona's stored face images to ground visual identity, and the model then retrieves one of five persona attributes: \emph{Biography} (text), \emph{Dialogue} (voice), \emph{Appearance} (image), \emph{Emotion} (voice), or \emph{Environment} (voice). All sub-tasks except \emph{Appearance} require a visual-to-text or visual-to-audio bridge.

    \item \emph{Group 2 (A2A; 5 tasks):} A voice query is matched against stored speaker samples to identify the persona, then sweeps the same five retrieval targets as Group 1. Sub-task 2-c (\emph{Appearance}) further requires an audio-to-visual bridge, a capability absent in purely unimodal retrieval.

    \item \emph{Group 3 (T2T; 4 tasks):} Identity is resolved through semantic matching between a textual query and the persona's stored textual profile, covering four retrieval targets: \emph{Biography} (text), \emph{Appearance} (image), \emph{Emotion} (text), and \emph{Environment} (text). Sub-task 3-b additionally crosses modality by retrieving a visual appearance description from a purely textual match. The \emph{Dialogue} target is intentionally omitted, as text-to-text conversational matching risks collapsing into shallow keyword overlap.

    \item \emph{Group 4 (T2Any; 4 tasks):} This group forms the benchmark's most demanding regime: a textual description of conversational content must be matched to the persona whose stored conversational audio semantically corresponds to it, without any explicit speaker cues. Retrieval targets span \emph{Biography} (text), \emph{Appearance} (image), \emph{Emotion} (voice), and \emph{Environment} (text). Sub-task 4-b requires a three-hop cross-modal path from the text query through audio-based identity matching to visual appearance retrieval, analogous to 2-c.
\end{itemize}

Across all four groups, every sub-task is paired with a no-GT (absent-persona) variant that requires the model to perform structured abstention, as summarized in Table~\ref{tab:task_taxonomy}. The elucidations of additional metrics used are in Table~\ref{tab:metric_summary}.
\begin{figure}[p]
  \centering
  \small
  \captionof{table}{Summary of evaluation metrics used in the Omni-Persona benchmark. All scores are reported with a higher-is-better polarity; MeanLen is descriptive and is excluded from Avg.}
  \label{tab:metric_summary}
  \renewcommand{\arraystretch}{1.0}
  \begin{tabularx}{\linewidth}{@{} l c >{\raggedright\arraybackslash}X @{}}
  \toprule
  \textbf{Metric} & \textbf{Target Subset} & \textbf{Description} \\
  \midrule
  \textbf{Overall}   & All & Item-weighted accuracy across both answerable and unanswerable scenarios. \\
  \textbf{Ans}       & Answerable & LLM-as-a-judge correctness rate. \\
  \textbf{Unans}     & Unanswerable & Successful abstention rate via keyword matching. \\
  \midrule
  \textbf{ROUGE-L}   & Answerable & Average LCS-based recall (sequence-sensitive lexical overlap). \\
  \textbf{Tok-F1}    & Answerable  & Average token-level F1 (bag-of-words lexical overlap). \\
  \textbf{BERTScore} & Answerable & Average BERTScore F1 (semantic embedding similarity). \\
  \midrule
  \textbf{Abs-F1}    & All & Harmonic mean of abstention precision and recall; penalizes over-abstention. \\
  \textbf{$1-\mathrm{FA}$} & Answerable & Complement of FalseAbs: fraction of answerable queries the model does \emph{not} answer with an abstention keyword. \\
  \textbf{TA}        & Unanswerable & TrueAbs: fraction of unanswerable queries on which the model correctly abstains. \\
  \midrule
  \textbf{Avg}       & Composite & Simple mean of the nine higher-is-better scores above (Overall, Ans, Unans, Abs-F1, ROUGE-L, Tok-F1, BERTScore, $1-\mathrm{FA}$, TA). \\
  \textbf{MeanLen}   & All & Mean response length in whitespace-tokenized words (descriptive; not part of Avg). \\
  \bottomrule
  \end{tabularx}

  \vspace{0em}

  \begin{tcolorbox}[
      colback=gray!5,
      colframe=gray!40,
      boxrule=0.5pt,
      arc=2pt,
      left=8pt, right=8pt, top=6pt, bottom=6pt,
      title=\textbf{Evaluation Metrics},
      fonttitle=\small\bfseries,
      coltitle=black,
      colbacktitle=gray!15,
  ]
  \small
  \textbf{Dual-Axis Recall}\\[2pt]
  We measure basic task completion via a dual-axis accuracy protocol that decouples successful answering from appropriate abstention.
  \begin{itemize}[leftmargin=*, itemsep=2pt, topsep=4pt]
      \item \textbf{Ans:} Proportion of the 391 answerable queries judged correct by an LLM-as-a-judge.
      \item \textbf{Unans:} Proportion of the 359 unanswerable queries for which the model triggers an abstention keyword (\textit{e.g.}, ``I can't decide'').
      \item \textbf{Overall:} Combined accuracy across all 750 items.
  \end{itemize}

  \vspace{0pt}\noindent\rule{\linewidth}{0.3pt}\vspace{0pt}

  \textbf{Generation Quality (Answerable Only)}\\[2pt]
  To assess the lexical and semantic fidelity of generated responses against the ground truth, we apply three established metrics on the answerable subset.
  \begin{itemize}[leftmargin=*, itemsep=2pt, topsep=4pt]
      \item \textbf{ROUGE-L Recall:} Structural overlap via the Longest Common Subsequence (LCS); strict on token order, penalizing valid paraphrasing.
      \item \textbf{Token F1:} Harmonic mean of token-level precision and recall under a bag-of-words representation, order-independent.
      \item \textbf{BERTScore F1:} Cosine similarity in the BERT embedding space, rewarding semantic equivalence beyond exact lexical matching.
  \end{itemize}

  \vspace{0pt}\noindent\rule{\linewidth}{0.3pt}\vspace{0pt}

  \textbf{Calibrated Abstention (Abs-F1).}
  Unanswerable recall alone is gameable by always abstaining. \textbf{Abs-F1} closes this loophole by treating unanswerable queries as the positive class: TP are correct abstentions, FP are erroneous abstentions on answerable items, and Abs-F1 is the harmonic mean of the resulting precision and recall. A high Abs-F1 thus reflects calibrated confidence, rejecting unanswerable queries without sacrificing valid answers.

  \vspace{0pt}\noindent\rule{\linewidth}{0.3pt}\vspace{0pt}

  \textbf{Anti-Hallucination Guardrails ($1-\mathrm{FA}$, $\mathrm{TA}$).}
  We report the two raw abstention rates jointly so that the \emph{direction} of a calibration failure is visible: $1-\mathrm{FA}$ drops when the model \emph{over-abstains} on answerable items (\textit{e.g.}, RL abstention-shortcut hacking), while $\mathrm{TA}$ drops when the model \emph{hallucinates} a grounded answer on unanswerable items. A model that maximizes one through blanket behavior is therefore automatically penalized by the other.

  \vspace{0pt}\noindent\rule{\linewidth}{0.3pt}\vspace{0pt}

  \textbf{Composite (Avg) and Length (MeanLen).}
  \textbf{Avg} is the simple mean of the nine higher-is-better scores listed above; it flags single-axis policies (e.g., universal abstention) that boost one metric while collapsing others. \textbf{MeanLen} is the mean response length in whitespace-tokenized words and is reported descriptively only, since it has neither a $[0,1]$ range nor a clear quality polarity.
  \end{tcolorbox}
\end{figure}

\begin{figure}[t!]
  \centering
\includegraphics[width=\columnwidth]{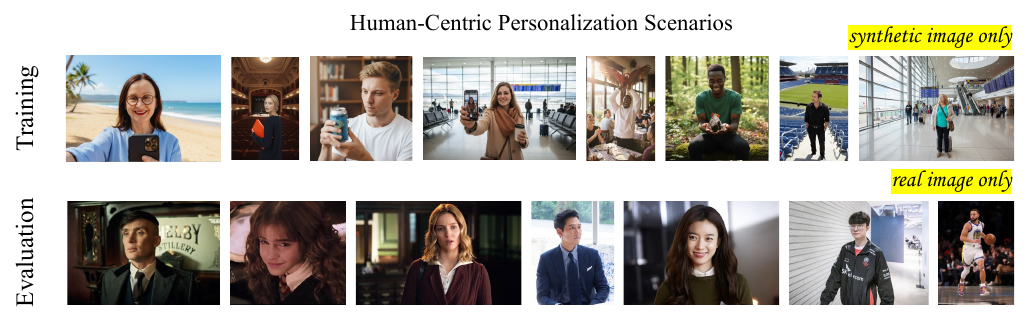}
\caption{\textbf{Qualitative image examples.} Sample images utilized during the training and evaluation phases. Note that synthetic images are strictly employed for SFT and RL training, whereas real images are reserved exclusively for the Omni-Persona benchmark.}\label{fig:scenario_query}
\end{figure}

\paragraph{Benchmark Complexity and Design Principles.} Our proposed benchmark is systematically more challenging in several respects.
The evaluation distractors are deliberately curated to be highly similar to the target, sharing specific vocal characteristics or visual resemblances, thereby requiring fine-grained cross-modal discrimination.
Moreover, each evaluation context contains four interleaved image-audio-text entries, exposing models to dense and heterogeneous multimodal signals.
Detailed per-scenario prompt templates and query-construction procedures are provided in Appendix~\ref{appendix:data_construction}.
Crucially, we introduce a deliberate distribution shift between training and evaluation to assess true generalization: training uses three context entries paired with synthesized CoViP images, whereas evaluation uses four context entries paired exclusively with real images.

\section{Evaluation Dataset Configurations}
\paragraph{Construction of Image--Text Pairs.}
To formulate diagnostic tasks for omnimodal personalization, it is essential to have multiple images of the same individual to facilitate cross-context identity reasoning. To this end, we sample person-related contexts from the evaluation split of CoViP, ensuring each individual appears as the query subject across diverse scenarios. This process yields 250 context--query pairs, each comprising a query image and four interleaved context entries, where each entry consists of an identity image paired with descriptive text.

To systematically construct the textual elements of the Omni-Persona benchmark, we refine the pre-generated dialogues through a multi-stage pipeline. First, a model extracts explicit personal attributes from the raw dialogue. It then plausibly imputes any missing traits and restructures the extracted information into concise personal profiles (\textit{e.g.}, in biography or dialogue format), strictly limited to 1--2 sentences per concept. By bifurcating scenarios into answerable and unanswerable (no-GT) cases, we synthesize targeted query prompts accordingly: for answerable cases, a GT answer is generated based on the target context; for unanswerable cases, the target is formatted to elicit structured abstention. All synthesis procedures are conducted using the proprietary frontier model \texttt{GPT-5.5}.

\paragraph{Audio Modality Construction.}
The audio modality consists of a balanced mixture of synthetic voice samples and real-world recordings, covering a total of 450 distinct speakers.

\textbf{Synthetic Data.}
For synthetic samples, we use \texttt{chatterbox}\footnote{\url{https://github.com/resemble-ai/chatterbox}} to generate high-fidelity audio (\textit{24-bit, 24\,kHz}). For each speaker, we construct pairs of distinct clips that share the same voice identity while varying emotion, conversational setting, and metadata such as age and accent.

\textbf{Real-world Data.}
Real-world conversational audio is curated from diverse corpora, including VoxMM~\citep{kwak2024voxmm}, MELD~\citep{poria2019meld}, JL-corpus~\citep{james2018open}, and RAVDESS~\citep{livingstone2018ryerson}. We extract 4--15 second segments and ensure expressive variation across two roles:
\begin{itemize}[leftmargin=*, noitemsep]
    \item \textbf{Reference Audio:} Uses relatively neutral or mild emotions, such as \textit{neutral}, \textit{calm}, and \textit{happy}.
    \item \textbf{Emotional Utterance:} Uses more expressive and distinct emotions, such as \textit{angry}, \textit{sad}, \textit{fearful}, \textit{disgust}, and \textit{surprised}.
\end{itemize}

\textbf{Context--Query Pair Construction.}
To construct challenging audio distractors for visually grounded pairs, we use \texttt{wav2vec2}~\citep{baevski2020wav2vec} for automated gender detection. Each target voice is then paired with a \textit{gender-aligned distractor} voice from either the synthetic or real-world subset. This setup prevents the model from relying on coarse gender cues and instead requires more fine-grained recognition of speaker-specific vocal characteristics.

In summary, the final Omni-Persona evaluation benchmark comprises approximately 750 items spanning 18 tasks across 4 scenario groups. This benchmark serves strictly as a held-out test set.


\begin{figure}[t!]
    \vspace{-1em}
    \centering

    \includegraphics[width=\columnwidth]
    {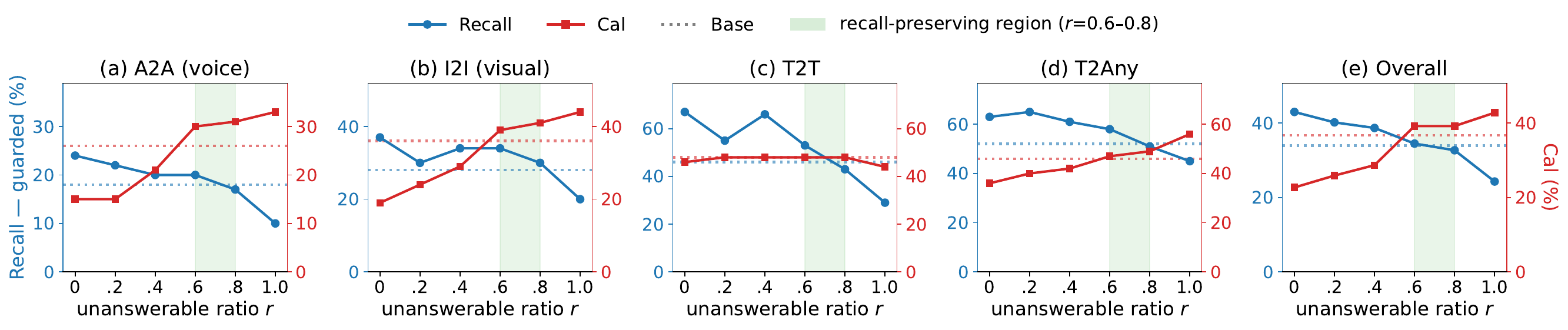}
    \caption{\textbf{Ablation of the answerable-to-unanswerable training ratio.} Varying the proportion of unanswerable examples reveals the trade-off between strict recall and calibration relative to the base model. Here, $r=0$ denotes training without unanswerable examples, consistent with the settings used in RePIC and CoViP.}

    \label{fig:sweet_spot}

    \vspace{0.5em}

    \includegraphics[width=\columnwidth]
    {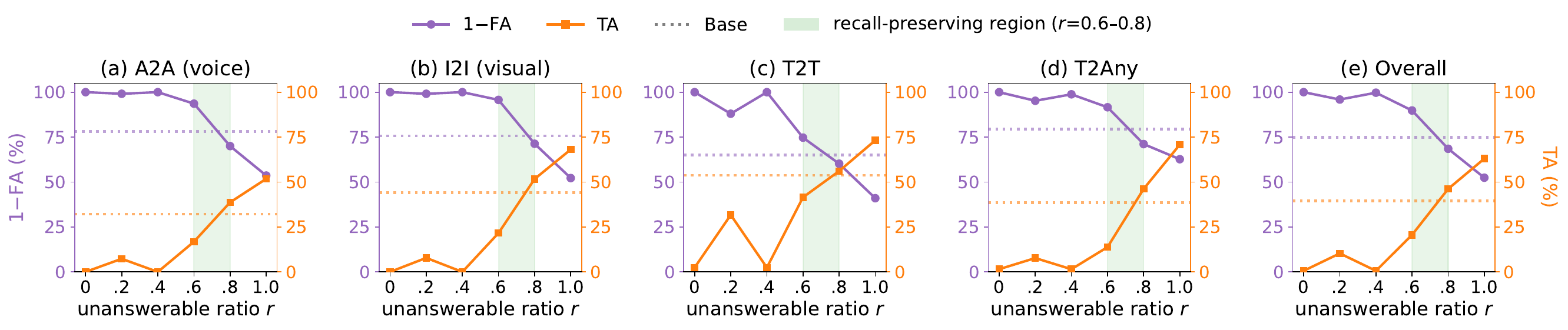}
    \caption{\textbf{Trade-off between $1-\mathrm{FA}$ and TA.} Varying the answerable-to-unanswerable ratio during post-training reveals a trade-off between correctly abstaining on unanswerable examples (high TA) and avoiding false abstention on answerable examples (high $1-\mathrm{FA}$). Here, $r=0$ denotes training without unanswerable examples, consistent with the settings of RePIC and CoViP.}

    \label{fig:sweet_spot_add}

    \vspace{-1em}
\end{figure}

\section{Additional Analysis}\label{sec:app_additional_analysis}
\vspace{-0.5em}

\subsection{Supervision-Induced Recall--Abstention Trade-off}
To assess the effect of labeled supervision composition during RLVR, we conduct all experiments using \texttt{Qwen2.5-Omni-3B} while varying the answerable-to-unanswerable training ratio from $1{:}0$ to $1{:}1$. As shown in Figures~\ref{fig:sweet_spot} and~\ref{fig:sweet_spot_add}, the guarded evaluation reveals a clear recall--abstention trade-off as the proportion of unanswerable examples increases. Answerable recall declines from 43.0 at a $1{:}0$ ratio to 24.3 at approximately $1{:}1$, while true-abstention accuracy rises from 0.6 to 63.0. This pattern suggests that abstention is primarily induced by explicit abstain-labeled supervision rather than emerging from grounding rewards or prompt-level instructions alone. In particular, the $1{:}0$ setting achieves the highest recall yet almost never abstains, demonstrating that strong grounding supervision is insufficient to produce calibrated abstention without explicit unanswerable examples.

A balanced 1:1 ratio achieves the highest aggregate calibration score but reduces recall below the base model by 9.7 points, showing that calibration is obtained at a substantial retrieval cost. In contrast, asymmetric supervision offers a more favorable operating point: the 1:0.6 setting restores recall to near-base performance while retaining substantial abstention accuracy, and ratios around 0.6--0.8 form a recall-preserving region on the empirical frontier. These results suggest that a relatively small but non-negligible amount of unanswerable supervision can mitigate the recall--abstention trade-off more effectively than fully balanced training. 
\subsection{Statistical Significance of the Recall--Calibration Trade-off}
\label{sec:app_significance}

We test whether the observed changes after RLVR reflect a consistent change in model behavior rather than random variation. Because the base and RLVR checkpoints are evaluated on the same items, we use McNemar's exact test for paired comparisons and report $95\%$ confidence intervals for each change. We conduct this analysis on the two \texttt{Qwen2.5-Omni} backbones for which paired item-level results are available.

As shown in Table~\ref{tab:app_significance}, RLVR significantly improves $\mathrm{Ans}$ and $1-\mathrm{FA}$ while significantly reducing $\mathrm{TA}$ on both backbones. This indicates that RLVR makes the model more likely to answer, improving recall on answerable queries but reducing appropriate abstention on unanswerable ones. Because these effects oppose each other, the resulting decrease in $\mathrm{Cal}$ is modest and not statistically significant for either backbone. Overall, the results support a clear shift toward answering, but not an improvement in calibrated behavior.

\begin{table}[h]
\centering
\small
\caption{\textbf{Paired significance of RLVR.} Comparison between base and RLVR checkpoints on the \texttt{Qwen2.5-Omni} backbones. Changes and $95\%$ confidence intervals are reported in percentage points, with significance assessed using McNemar's exact test.}
\label{tab:app_significance}
\setlength{\tabcolsep}{6pt}
\begin{tabular}{@{}lccccc@{}}
\toprule
\textbf{Metric} & \textbf{Base} & \textbf{RLVR} & $\Delta$ & \textbf{95\% CI} & $p$ \\
\midrule
\multicolumn{6}{@{}l}{\textit{Qwen2.5-Omni-3B}} \\
\quad Ans  & 34.0 & 40.9 & $+6.9$  & [$+2.8$, $+11.0$] & 0.002 \\
\quad Cal  & 36.7 & 34.1 & $-2.6$  & [$-5.3$, $+0.3$]  & 0.093 \\
\quad 1-FA & 74.9 & 84.7 & $+9.8$  & [$+6.5$, $+12.9$] & $<10^{-8}$ \\
\quad TA   & 39.6 & 26.7 & $-12.9$ & [$-16.3$, $-9.4$] & $<10^{-13}$ \\
\midrule
\multicolumn{6}{@{}l}{\textit{Qwen2.5-Omni-7B}} \\
\quad Ans  & 38.6 & 47.8 & $+9.2$  & [$+4.3$, $+14.1$] & $<10^{-3}$ \\
\quad Cal  & 30.9 & 28.4 & $-2.5$  & [$-5.8$, $+0.7$]  & 0.151 \\
\quad 1-FA & 82.1 & 98.0 & $+15.9$ & [$+12.2$, $+19.5$] & $<10^{-17}$ \\
\quad TA   & 22.6 & 7.2  & $-15.4$ & [$-19.2$, $-11.4$] & $<10^{-14}$ \\
\bottomrule
\end{tabular}
\end{table}

\subsection{Reward Asymmetry on the Recall--Abstention Trade-off}
\label{sec:app_reward_asymmetry}
We examine whether the recall--abstention trade-off can be adjusted through asymmetric reward assignments. All experiments use \texttt{Qwen2.5-Omni-3B} with a fixed answerable-to-unanswerable training ratio of $1{:}0.6$. The training data, optimization procedure, and all other configurations are held constant; only the rewards assigned to different response outcomes are varied. 

For an answerable query, a correct response is treated as a true positive (TP), whereas abstention is a false negative (FN), corresponding to false abstention. For an unanswerable query, abstention is a true negative (TN), whereas providing an unsupported answer is a false positive (FP), corresponding to false identification. Correct answers receive a fixed reward of $+1.0$ in all configurations, while the rewards or penalties for FN, TN, and FP outcomes are varied as summarized in Table~\ref{tab:app_reward_asymmetry_results}.

\paragraph{Asymmetric Reward Ablation.}
The reward configurations are designed to distinguish improved answerability discrimination from a general tendency to answer more frequently. Comparing \emph{Abstention-softened} with \emph{Symmetric} isolates the effect of reducing both the reward for correct abstention and the penalty for false abstention, while keeping the false-identification penalty fixed. Comparing \emph{Error-softened (Ours)} with \emph{Symmetric} instead halves both error penalties while retaining the full reward for correct abstention.

A symmetric assignment penalizes false abstention as strongly as false identification, which limits systematic bias toward either answering or abstaining but also suppresses answering overall. Our final configuration therefore softens both error penalties while retaining the full reward for correct abstention. The ablation evaluates whether this asymmetric shaping provides a more favorable balance between answering supported queries and abstaining from unsupported ones.

\begin{table}[t]
\centering
\small
\setlength{\tabcolsep}{6pt}
\caption{
\textbf{Reward asymmetry ablation.} Left: reward configurations used to separate the
effects of abstention-related rewards from the penalty for unsupported answers; the TP
reward is fixed at $+1.0$ in all configurations. Right: the corresponding effect on
answerable performance, abstention behavior, and calibrated accuracy.
}
\label{tab:app_reward_asymmetry_results}
\begin{tabular}{l ccc | cccc}
\toprule
& \multicolumn{3}{c|}{\textbf{Reward configuration}}
& \multicolumn{4}{c}{\textbf{Result}} \\
\cmidrule(lr){2-4}\cmidrule(l){5-8}
Configuration & TN & FN & FP
& Recall $\uparrow$
& $1-\mathrm{FA}$ $\uparrow$
& TA $\uparrow$
& Cal. $\uparrow$ \\
\midrule
Base                 & --     & --     & --     & 34.0 & 74.9 & \textbf{39.6} & \textbf{36.7} \\
Symmetric           & $+1.0$ & $-1.0$ & $-1.0$ & 30.9 & 89.2 & 28.8 & 30.1 \\
Abstention-softened  & $+0.5$ & $-0.5$ & $-1.0$ & 39.1 & \textbf{95.7} & 8.4  & 24.4 \\
Error-softened (Ours)      & $+1.0$ & $-0.5$ & $-0.5$ & \textbf{40.9} & 84.7 & 26.7 & 34.1 \\
\bottomrule
\end{tabular}
\end{table}
Table~\ref{tab:app_reward_asymmetry_results} shows that all post-training configurations shift the model toward answering more frequently. Relative to the base model, the symmetric configuration raises $1-\mathrm{FA}$ from $74.9\%$ to $89.2\%$ while TA decreases from $39.6\%$ to $28.8\%$, and recall drops from $34.0\%$ to $30.9\%$: penalizing both error types equally increases answering attempts without converting them into correct answers.

Softening the abstention-related rewards further amplifies this tendency. Under \emph{Abstention-softened}, $1-\mathrm{FA}$ rises to $95.7\%$, while TA falls to $8.4\%$. This change therefore reflects a general decrease in abstention rather than improved discrimination between answerable and unanswerable inputs. In contrast, our \emph{Error-softened} configuration, which halves both error penalties while keeping the full abstention reward, attains the highest recall ($40.9\%$) while retaining TA at $26.7\%$ and the best calibrated accuracy among the post-training configurations ($34.1\%$).

Overall, how the softening is distributed matters. Softening only the abstention-related rewards trades TA away for a higher attempt rate, whereas softening both error penalties equally, as in our configuration, yields the best recall--calibration balance among the post-training configurations. Even so, its calibrated accuracy remains below the base model, indicating that reward asymmetry alone does not fully resolve the recall--abstention trade-off.

Compared with a symmetric non-negative reward, this formulation assigns a penalty to false negatives and a larger reward to correctly recognized positive matches. It therefore discourages a policy from improving its apparent reliability by indiscriminately rejecting candidate identities. False positives remain explicitly penalized to expose the opposite failure mode of indiscriminate positive prediction.

\subsection{Reward-Component Ablation}
\label{sec:app_reward_components}

We next isolate the contribution of each verifiable reward component. Starting from the full
reward (localization, verification, and text QA; Appendix~\ref{appendix:training_details}), we
remove one component at a time and additionally train a text-QA-only variant. All runs use
\texttt{Qwen2.5-Omni-3B} under identical settings.

\begin{table}[t]
\centering
\small
\setlength{\tabcolsep}{6pt}
\caption{
\textbf{Reward-component ablation.} Localization drives answerable recall, whereas removing it collapses attempts into
abstention and inflates $\mathrm{Cal}$.
}
\label{tab:app_reward_components}
\begin{tabular}{l cccc | cc}
\toprule
& \multicolumn{4}{c|}{\textbf{Overall}}
& \multicolumn{2}{c}{\textbf{Perceptual Ans}} \\
\cmidrule(lr){2-5}\cmidrule(l){6-7}
Configuration
& Recall $\uparrow$
& $1-\mathrm{FA}$ $\uparrow$
& TA $\uparrow$
& Cal. $\uparrow$
& I2I & A2A \\
\midrule
Base (no RLVR)      & 34.0 & 74.9 & 39.6 & 36.7 & 27.8 & 18.2 \\
Full (Ours)         & \textbf{40.7} & \textbf{84.9} & 26.7 & 34.0 & \textbf{32.2} & \textbf{26.4} \\
\quad w/o Verification  & 37.6 & 74.4 & 39.0 & 35.3 & 26.6 & 20.5 \\
\quad w/o Localization  & 31.5 & 60.9 & \textbf{61.3} & \textbf{45.7} & 21.7 & 18.2 \\
\quad Text-QA only      & 31.4 & 73.4 & 43.5 & 32.7 & 22.1 & 12.7 \\
\bottomrule
\end{tabular}
\end{table}

Table~\ref{tab:app_reward_components} shows that the reward components jointly govern the trade-off between recall and abstention. \emph{Localization} is the primary driver of answerable recall: removing it reduces overall recall from $40.7\%$ to $31.5\%$, below the base model, while I2I recall falls from $32.2\%$ to $21.7\%$ and A2A recall returns to the base level ($26.4\% \rightarrow 18.2\%$). At the same time, $\mathrm{TA}$ rises sharply from $26.7\%$ to $61.3\%$, while $1-\mathrm{FA}$ falls from $84.9\%$ to $60.9\%$, indicating a strong shift toward abstention on both unanswerable and answerable examples. This shift yields the highest $\mathrm{Cal}$ score ($45.7\%$), but the gain reflects increased abstention rather than improved perceptual grounding.

Removing \emph{verification} produces a more moderate decline, reducing overall recall by $3.1$ points and perceptual recall by $5.6$ points on I2I and $5.9$ points on A2A. Although $\mathrm{TA}$ returns close to the base level ($39.0\%$ versus $39.6\%$), $\mathrm{Cal}$ remains below the base model ($35.3\%$ versus $36.7\%$). The \emph{text-QA-only} variant performs worse than the base model in overall recall ($31.4\%$ versus $34.0\%$), $\mathrm{Cal}$ ($32.7\%$ versus $36.7\%$), and perceptual recall (I2I $22.1\%$ versus $27.8\%$; A2A $12.7\%$ versus $18.2\%$), demonstrating that text-only supervision does not account for the gains of the full reward.

Overall, the full reward achieves the highest overall and perceptual recall, whereas removing localization yields the highest $\mathrm{Cal}$ by substantially increasing abstention. These results identify localization as the main source of the recall--abstention tension and verification as an important contributor to perceptual grounding. They also show that the reward components are complementary: removing either component weakens recall, while text-QA supervision alone fails to improve either recall or calibration over the base model.

\subsection{Additional Cross-Judge Robustness}
\label{sec:app_additional_cross_judge}

To further assess the robustness of the LLM-based evaluation, we re-evaluated all answerable instances using GPT-5.6-sol with medium reasoning effort. The system prompt and evaluation procedure were identical to those used for the primary GPT-5.4-mini judge, with only the judge model changed. This check was run on an earlier snapshot of the model pool ($11$ models, $391$ answerable instances each, no API failures), so it probes the reliability of the judge rather than re-scoring the final checkpoints reported in Table~\ref{tab:main_results}. Unanswerable instances remained keyword-scored and were therefore unaffected by the choice of LLM judge.

GPT-5.6-sol showed item-level agreement of $80.6\text{--}92.1\%$ with the primary judge, with Cohen's $\kappa$ ranging from $0.61$ to $0.81$. Although GPT-5.6-sol assigned lower answerable scores to most models, the overall model ordering remained similar: the Spearman rank correlation was $\rho=0.91$ for calibrated accuracy ($p < 0.001$) and $\rho=0.66$ for answerable recall ($p = 0.026$). Changing the judge therefore shifts absolute answerable scores more than it shifts the calibrated-accuracy ranking.

Table~\ref{tab:app_cross_judge_summary} summarizes the results across the two alternative judges. In both cases, the calibrated-accuracy ranking was strongly correlated with that obtained using the primary judge, while the answerable-recall ranking was more judge-sensitive.

\begin{table}[t]
\centering
\small
\setlength{\tabcolsep}{5pt}
\caption{
Judge robustness under alternative LLM judges, measured on an earlier snapshot of the
model pool ($11$ models). Rank correlations are computed against the primary GPT-5.4-mini judge.
}
\label{tab:app_cross_judge_summary}
\begin{tabular}{lccc}
\toprule
Alternative judge
& Cal. rank correlation
& Ans. rank correlation
& Top model preserved \\
\midrule
Gemini-3.1-Pro
& $\rho=0.89$ ($p < 0.001$)
& $\rho=0.63$ ($p = 0.039$)
& Yes \\
GPT-5.6-sol
& $\rho=0.91$ ($p < 0.001$)
& $\rho=0.66$ ($p = 0.026$)
& Yes \\
\bottomrule
\end{tabular}
\end{table}

Overall, the consistency across two alternative judges indicates that the principal calibrated-accuracy ranking is reasonably robust to the choice of evaluator.

\begin{figure*}[t]
    \centering
    \includegraphics[width=\textwidth]{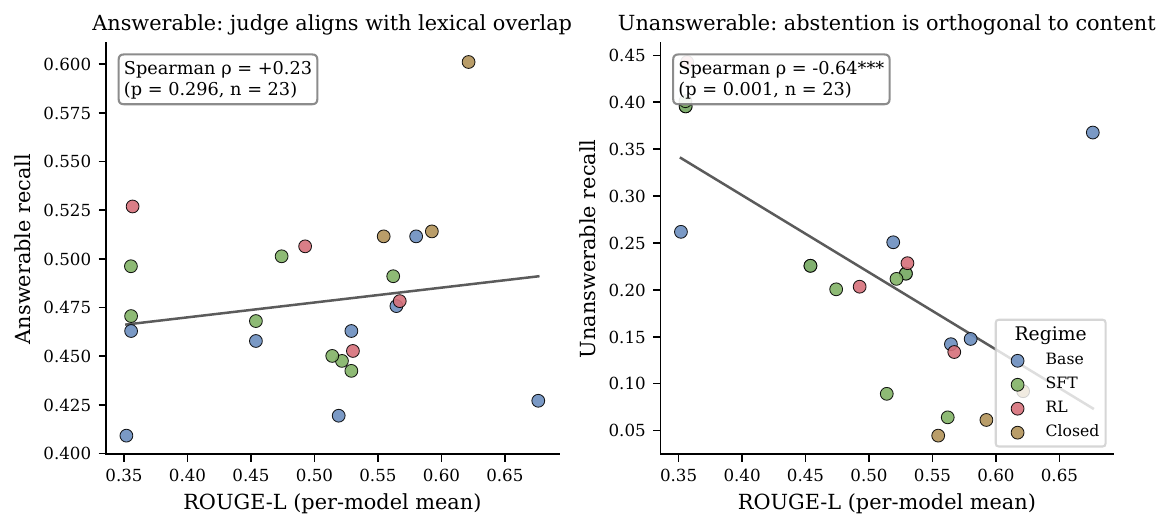}
    \caption{Metric alignment across post-training regimes. The plot compares judge scores and lexical overlap for answerable and unanswerable queries. Crucially, it demonstrates the absence of content-induced bias in unanswerable cases, showing that abstention behavior is independent of content overlap.}    \label{fig:metric_1}
\end{figure*}

\begin{figure*}[!t]
    \centering
    \includegraphics[width=0.85\textwidth]{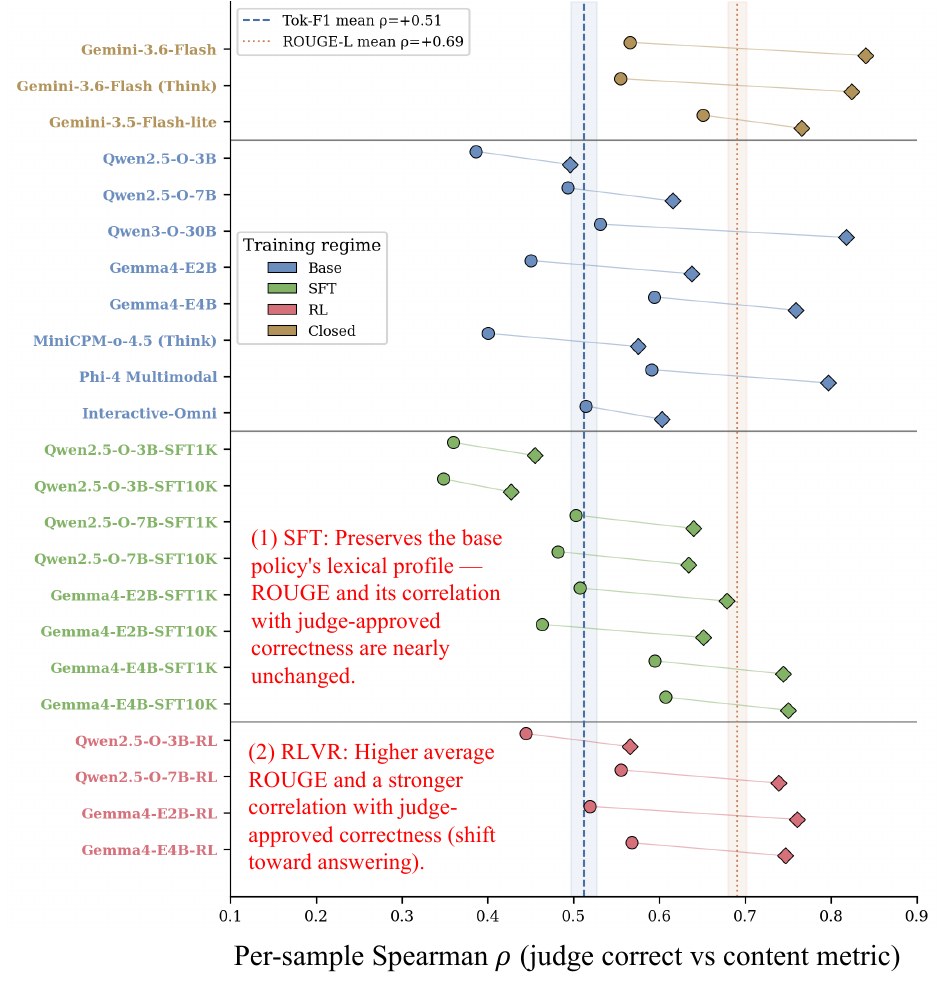}
    \caption{Per-model agreement on answerable items only between judge and content metrics (circle = \texttt{Tok-F1}, diamond = \texttt{ROUGE-L})}
    \label{fig:metric_2}
\end{figure*}


\subsection{Alignment of Evaluation Metrics}\label{appendix:judge_reliability}
LLM-as-a-judge evaluations and traditional lexical/semantic metrics capture fundamentally distinct signals.
Because their correspondence varies substantially across query types, this benchmark reports answerable QA quality and unanswerable abstention behavior separately.

\textbf{Why LLM-as-a-judge is Needed to Evaluate Free-Form Responses:} 
Figure~\ref{fig:metric_1} demonstrates that content-overlap metrics primarily explain performance on answerable queries, but fail to capture abstention behavior on unanswerable queries. That is, content quality and abstention capability are distinct evaluation axes. While answerable recall increases with ROUGE-L, indicating that lexical overlap partially reflects answer quality, unanswerable recall shows little association with it. This suggests that abstention behavior cannot be inferred from traditional content metrics and must be evaluated separately. 

\begin{keyfinding}{Key Finding 4: RLVR Improves Lexical Overlap and Judge--Metric Agreement}
Unlike SFT, which largely preserves the lexical profile of the base policy, RLVR shifts model outputs toward answering. This raises conventional lexical overlap (ROUGE-L $+9.2$/$+10.0$ points on the Qwen backbones and modestly on the Gemma4 models) and, together with it, strengthens the per-sample agreement between the LLM judge and lexical metrics.
\end{keyfinding}

\textbf{Per-Sample Metric Reliability:} As illustrated in Figure~\ref{fig:metric_2} and Table~\ref{tab:main_results_refined}, ROUGE-L exhibits stronger agreement with LLM-as-a-judge evaluations than token-level F1 on answerable items across all models. Strong closed-source and base models maintain high concordance between the judge and lexical metrics, and RL-tuned models attain the strongest alignment among the post-trained regimes (mean $\rho=0.70$ versus $0.63$ for the base policies). This indicates that reward optimization moves outputs toward grounded answering, which both raises corpus-level lexical overlap and makes lexical metrics track judge verdicts more closely. Conversely, SFT models largely preserve the judge--metric alignment of their base counterparts (mean $\rho=0.62$), suggesting that supervised fine-tuning induces little lexical deviation from the original policy. 


\section{Additional Experimental Configurations}
\label{appendix:training_details}

\subsection{Details on SFT Implementations}

\paragraph{SFT Training Objective.}
We train the model on $\mathcal{D}_{\mathrm{SFT}}$, a dataset of triplets $(q, \mathcal{C}, y^*)$, where $q$ is a query, $\mathcal{C}$ is the omnimodal context, and $y^*$ is the ground-truth response. We minimize the standard autoregressive negative log-likelihood:
\begin{equation}
\mathcal{L}_{\mathrm{SFT}}(\theta) = - \mathbb{E}_{(q, \mathcal{C}, y^*) \sim \mathcal{D}_{\mathrm{SFT}}} \sum_{t=1}^{|y^*|} \log \pi_\theta(y^*_t \mid q, \mathcal{C}, y^*_{<t}).
\label{eq:sft_obj}
\end{equation}

\paragraph{SFT Configurations.} We fine-tune four backbones via ms-swift: \texttt{Qwen2.5-Omni} (3B/7B) and \texttt{Gemma4} (E2B/E4B). All SFT runs use $\mathrm{lr}=2\times10^{-5}$ over 3 epochs with vision/audio encoders frozen. We apply LoRA ($r=64, \alpha=128$) for the Qwen backbones and a more compact LoRA ($r=32, \alpha=64$) for the Gemma4 models. All post-training is performed with LoRA~\citep{hu2022lora}, using ms-swift~\citep{msswift} for SFT. Within the Gemma4 series, audio processing is supported exclusively by the E2B and E4B variants.

\paragraph{SFT Data Corpus Construction.}
\begin{figure}[p]
  \centering
  \vspace{-1em}\includegraphics[width=\columnwidth]{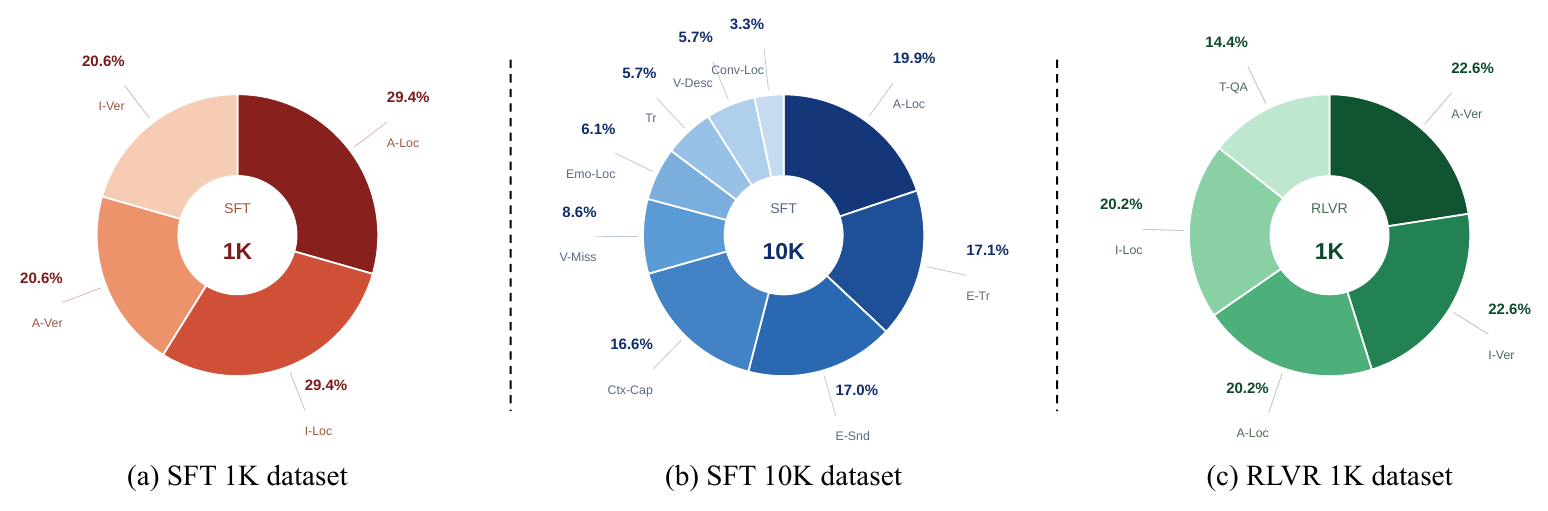}
  \caption{\textbf{Post-training data strategy.} Visualization of the task coverage and data distributions across the SFT (1K), SFT (10K), and RLVR training regimes.}
  \label{fig:task_distribution}
  \vspace{0.6em}
  \centering
\captionof{table}{\textbf{SFT (1K) task distribution}. Overview of the task proportions and corresponding abbreviations utilized in our post-training ablations.}
\label{tab:localize_verify_task_abbreviations}
\resizebox{\linewidth}{!}{%
\begin{tabular}{@{}ll@{}}
\toprule
\textbf{Task name} & \textbf{Sample query text} \\
\midrule
Audio localization & Select which reference audio panel contains the same speaker as the query audio. \\
Image localization & Select which reference image panel contains the same person/object as the query image. \\
Audio verification & Answer yes/no on whether the query audio matches the referenced speaker/audio identity. \\
Image verification & Answer yes/no on whether the query image matches the referenced person/object identity. \\
\bottomrule
\end{tabular}%
}

  \vspace{0.4em}
  \centering
\captionof{table}{\textbf{SFT (10K) task distribution}. Overview of the task proportions and corresponding abbreviations utilized in our post-training ablations.}
\label{tab:sft_task_abbreviations}
\resizebox{\linewidth}{!}{%
\begin{tabular}{@{}ll@{}}
\toprule
\textbf{Task name} & \textbf{Sample query text} \\
\midrule
Audio localization & Select the reference speaker/audio panel that matches the query audio. \\
Event-aware transcription & Transcribe speech while handling an added environmental sound event. \\
Event sound MCQA & Identify the environmental sound mixed with speech from multiple choices. \\
Contextualized captioning & Describe the query image using the matching memory/context panel when relevant. \\
Missing-image visual description & Describe visual information when the target image modality is absent. \\
Emotion speaker localization & Find which reference speaker matches the query speech/emotion cue. \\
Speech transcription & Convert the target speech audio into text. \\
Visual description & Generate a grounded description of the target image. \\
Conversation speaker localization & Identify the matching speaker/person from dialogue-conditioned references. \\
\bottomrule
\end{tabular}%
}

  \vspace{0.4em}
  \centering
\captionof{table}{\textbf{Task distribution in the RLVR (1K) dataset.} This table summarizes the task abbreviations used in the RLVR training distribution. For all referential tasks, the target responses are restricted to binary ``Yes'' or ``No'' labels.}
\label{tab:rlvr_task_abbreviations}
\resizebox{\linewidth}{!}{%
\begin{tabular}{@{}ll@{}}
\toprule
\textbf{Task name} & \textbf{Sample query text} \\
\midrule
Audio verification & Decide whether the query audio matches the referenced speaker/audio identity. \\
Image verification & Decide whether the query image matches the referenced person/object identity. \\
Audio localization & Select which reference audio panel contains the same speaker as the query audio. \\
Image localization & Select which reference image panel contains the same person/object as the query image. \\
Text QA & Answer a context-grounded question or abstain when the answer is not supported. \\
\bottomrule
\end{tabular}%
}

  \vspace{0.4em}
  \centering
\small
\captionof{table}{Overview of data configurations for training and evaluation phases.}
\label{tab:data_overview}
\setlength{\tabcolsep}{4pt}
\renewcommand{\arraystretch}{1.3} 

\begin{tabularx}{\linewidth}{@{} l c c 
    >{\raggedright\arraybackslash\hsize=1.45\hsize}X 
    >{\raggedright\arraybackslash\hsize=0.85\hsize}X 
    >{\raggedright\arraybackslash\hsize=0.85\hsize}X 
    >{\raggedright\arraybackslash\hsize=0.85\hsize}X @{}}
\toprule

\multirow{2}{*}{\textbf{Stage}} 
& \multirow{2}{*}{\textbf{Size}} 
& \multirow{2}{*}{\textbf{\# of Contexts}} 
& \multirow{2}{*}{\textbf{Tasks}} 
& \multicolumn{3}{c}{\textbf{Source Modality}} \\
\cmidrule(l){5-7}
& & & & \textbf{Image} & \textbf{Text} & \textbf{Audio} \\
\midrule

SFT Corpus 
& 1K / 10K
& 3
& 12 foundational task types
& Synthetic-only
& Synthetic (\texttt{GPT-5.4}) 
& Real \& Syn. (80\%) \\

RLVR Corpus
& 1K
& 3
& 3 core grounding tasks (Visual ID, Voice ID, Text QA)
& Synthetic-only
& Synthetic (\texttt{GPT-5.4}) 
& Real \& Syn. (80\%) \\

Eval Benchmark 
& 750  
& 4
& 18 fine-grained tasks (4 groups) 
& Real-only
& Synthetic (\texttt{GPT-5.5})
& Real \& Syn. (20\%) \\

\bottomrule
\end{tabularx}
\end{figure}

The 10K-scale SFT corpus (detailed in Table~\ref{tab:sft_task_abbreviations}) encompasses 12 distinct task types explicitly designed to enforce robust omnimodal alignment. The dataset comprehensively integrates foundational grounding tasks, audio-centric scenarios (\textit{e.g.}, conversational and event sounds), and missing-modality variants that train the model to recognize and verbalize when required context is absent. To mitigate positional and cross-modal shortcut biases, we apply rigorous augmentations such as context reordering, distractor replacement, and modality swapping. Contextual memory is sourced from CoViP dialogues~\citep{covip} and CUPID biographies~\citep{kim2025cupid}, with ground-truth visual captions rigorously filtered by an MCQA accuracy threshold ($\ge$50\%). A downscaled 1K subset is also provided for experimental efficiency (Table~\ref{tab:localize_verify_task_abbreviations}).

\paragraph{Training Data Corpus Overview.}
Our experimental pipeline comprises three distinct data stages (see Table~\ref{tab:data_overview}). The SFT corpus (1K and 10K variants) provides broad modality-alignment supervision, examining the impact of data scaling. Subsequently, the RLVR corpus supplies VR signals focused on two core grounding tasks: audio and image perception, and text-grounded retrieval. Notably, across both the SFT and RLVR regimes, we ensure that absent-persona (no-GT) samples account for approximately $37.5\%$ of the training mixtures, following the $1{:}0.6$ answerable-to-unanswerable ratio. This deliberate inclusion is intended to encourage calibrated abstention when confronted with unanswerable queries. The resulting trade-off between answerable recall and Cal is governed jointly by the reward design and the answerable-to-unanswerable data composition, as analyzed through the ratio sweep (Figures~\ref{fig:sweet_spot} and~\ref{fig:sweet_spot_add}) and the reward-asymmetry ablation (Appendix~\ref{sec:app_reward_asymmetry}). The dataset distribution across each training stage is illustrated in Figure~\ref{fig:task_distribution}.

\begin{figure}[t!]
  \centering
  \includegraphics[width=\columnwidth]{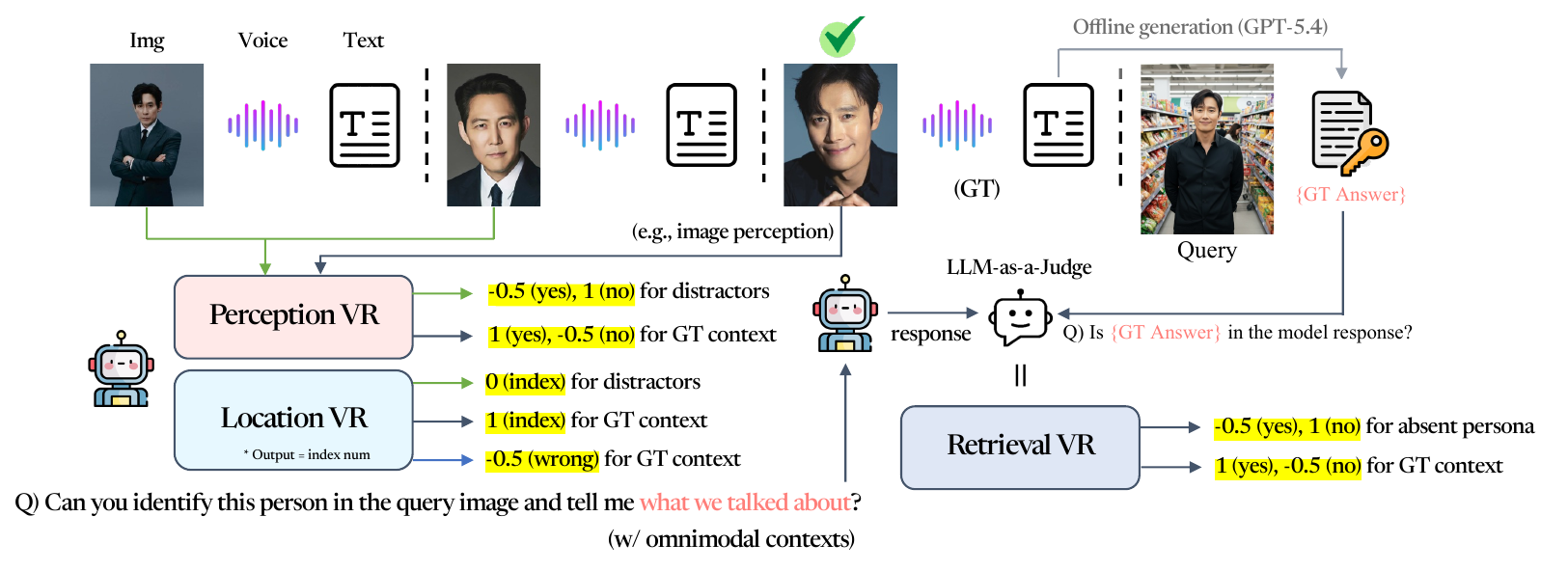}
\caption{\textbf{Overview of the perception, localization, and retrieval rewards used for RLVR.} The rewards use error-softened scoring rather than strictly binary signals. Perception and localization rewards supervise identification of the target context and its position among distractors, while the retrieval reward evaluates grounded answering when the target persona is present and appropriate abstention otherwise. Answer correctness is assessed against offline reference answers using an LLM judge. Training-data details are provided in Table~\ref{tab:data_overview}.}
\label{fig:rlvr_overview}

    \vspace{-1em}
\end{figure}

\subsection{Details on RLVR Implementations}
\label{subsec:algo_choice}
We adopt Group Sequence Policy Optimization (GSPO;
\citealp{zheng2025group}) as the base optimizer for our RLVR, with a moderate
KL regularization coefficient $\beta = 0.04$. 

\paragraph{RLVR Training Objective.}
We optimize $\pi_\theta$ to maximize a verifiable reward while penalizing the KL divergence~\citep{shao2024deepseekmath, zheng2025group} from the reference policy $\pi_{\mathrm{ref}}$:
\begin{equation}
    \max_{\theta} \mathbb{E}_{(q, \mathcal{C}) \sim \mathcal{D}_{\mathrm{tr}}} \mathbb{E}_{y \sim \pi_{\theta}(\cdot \mid q, \mathcal{C})} \left[ r(y, q, \mathcal{C}) - \beta \mathrm{D}_{\mathrm{KL}} \left( \pi_{\theta}(\cdot \mid q, \mathcal{C}) \| \pi_{\mathrm{ref}}(\cdot \mid q, \mathcal{C}) \right) \right]
\end{equation}
here, $q$ is a query, $\mathcal{C}$ is omnimodal contexts, $y$ is the model response. For each training instance, the verifiable reward $r(y, q, \mathcal{C})$ is specifically determined based on the success of $y$ in the assigned task as follows:

\paragraph{RLVR Configurations.} Without an SFT warmup stage, we optimize the RLVR objective using GSPO~\citep{zheng2025group} on 1K samples. Key hyperparameters include: $G=4$ generations, $\beta=0.04$, $\mathrm{lr}=1\times10^{-5}$, and LoRA $r=32, \alpha=64$. Similar to SFT, we freeze all perceptual encoders to maintain pre-trained representations. We note that our post-training hyperparameters were tuned primarily on \texttt{Qwen2.5-Omni-3B} and reused across backbones, so they may be suboptimal for the Gemma4 models.

\textbf{Verifiable Reward Designs.}
Mirroring the three reward components in Section~\ref{sec:targeted_rl}, our reward function is
dispatched by task type $t(q)$ into \emph{perception}, \emph{location}, and \emph{retrieval}:
\begin{equation}
r_{\mathrm{base}} =
\begin{cases}
r_{\mathrm{perc}}(y, q, \mathcal{C}), & \text{if } t(q) = \textsc{perc}, \\
r_{\mathrm{loc}}(y, q, \mathcal{C}), & \text{if } t(q) = \textsc{loc}, \\
r_{\mathrm{qa}}(y, q, \mathcal{C}), & \text{if } t(q) = \textsc{qa}.
\end{cases}
\label{eq:reward_base_app}
\end{equation}
All three share the same \emph{asymmetric} outcome scheme. Writing the four outcomes of a
verifiable decision as true positive (TP; a correct grounded answer or a correct match), true
negative (TN; a correct abstention or a correct rejection), false negative (FN; false abstention
or a missed match), and false positive (FP; an unsupported answer or a false match), we assign
\begin{equation}
r(\text{TP}) = r(\text{TN}) = +1.0, \qquad
r(\text{FN}) = r(\text{FP}) = -0.5 ,
\label{eq:reward_asym_app}
\end{equation}
with all remaining outcomes (\textit{e.g.}, an incorrect grounded answer or an unparsable
response) scored $0$. Rewarding TP and TN equally prevents the policy from inflating its apparent
reliability through blanket answering or blanket abstention, while the softened $-0.5$ penalties
keep both error types costly without over-suppressing attempts; this asymmetry is ablated in
Appendix~\ref{sec:app_reward_asymmetry}.

\textbf{\textit{1) Perception VR.}}
Perception VR supervises whether a target perceptual signal, such as an image or audio clip, matches a given context. Queries expose only the target modality to prevent cross-modal shortcutting. Correct matches and rejections receive $+1.0$, while false negatives and false positives receive $-0.5$.

\textbf{\textit{2) Location VR.}}
Location VR provides finer-grained supervision by requiring the model to select the matching context index or a ``None'' option when no match exists. Correct index selection and correct abstention receive $+1.0$; false negatives and false positives receive $-0.5$; incorrect indices and unparsable outputs receive $0$.

\textbf{\textit{3) Retrieval VR.}}
Retrieval VR supervises grounded open-ended answering and appropriate abstention. When the target persona is present, an LLM judge evaluates answer correctness; when absent, abstention is checked using the evaluation-time lexical rule. Correct answers and abstentions receive $+1.0$, false negatives and false positives receive $-0.5$, and incorrect grounded answers receive $0$. Although these rewards explicitly supervise abstention, the resulting policies still exhibit a bias toward answering.

\textbf{Safeguards Against Reward Hacking.}
We include several safeguards to reduce reward hacking during RL fine-tuning. Outputs with clear degeneration patterns, such as repeated 4-grams, character-level spam, or sentence loops, receive zero reward. We also parse LLM-judge outputs with strict exact matching, so that only the verdict \texttt{correct} is rewarded and strings such as \texttt{incorrect} cannot be mistakenly counted as positive. Invalid ground-truth labels are treated as explicit errors rather than ignored silently. These safeguards make the reward signal more reliable while keeping the training procedure simple.

\paragraph{On-Policy RLVR Algorithms.}
Given a prompt $x$, group size $G$, and sampled responses $\{y_i\}_{i=1}^{G}$
with group-normalized advantages $\hat{A}_i$, the two objectives differ in the
granularity at which the importance ratio is computed and clipped:
\begin{equation}
\mathcal{J}_{\mathrm{GRPO}}(\theta)
= \mathbb{E}\!\left[
\frac{1}{G}\sum_{i=1}^{G}\frac{1}{|y_i|}\sum_{t=1}^{|y_i|}
\min\!\left(
r_{i,t}(\theta)\,\hat{A}_i,\;
\mathrm{clip}\!\left(r_{i,t}(\theta),\, 1{\pm}\varepsilon\right)\hat{A}_i
\right)
\right]
- \beta\,\mathbb{D}_{\mathrm{KL}}\!\left[\pi_\theta \,\|\, \pi_{\mathrm{ref}}\right],
\label{eq:grpo}
\end{equation}
\begin{equation}
\mathcal{J}_{\mathrm{GSPO}}(\theta)
= \mathbb{E}\!\left[
\frac{1}{G}\sum_{i=1}^{G}
\min\!\left(
s_i(\theta)\,\hat{A}_i,\;
\mathrm{clip}\!\left(s_i(\theta),\, 1{\pm}\varepsilon\right)\hat{A}_i
\right)
\right],
\label{eq:gspo}
\end{equation}
where $r_{i,t}(\theta) = \pi_\theta(y_{i,t}\mid x, y_{i,<t})\,/\,
\pi_{\mathrm{old}}(y_{i,t}\mid x, y_{i,<t})$ is the per-token ratio, and
\begin{equation}
s_i(\theta) \;=\; \left(\frac{\pi_\theta(y_i \mid x)}{\pi_{\mathrm{old}}(y_i \mid x)}\right)^{\!1/|y_i|}
\label{eq:seq_ratio}
\end{equation}
is the length-normalized sequence-level ratio. Both objectives share the same
KL reference $\pi_{\mathrm{ref}}$; the substantive distinction lies in
Eq.~\eqref{eq:seq_ratio} and in whether clipping is applied per-token
(Eq.~\eqref{eq:grpo}) or per-sequence (Eq.~\eqref{eq:gspo}).

\paragraph{Why Sequence-Level Optimization Suits our RLVR.}
Our RLVR verifier combines rule-based binary verification (e.g., 'yes'/'no' compliance) with an LLM-as-judge component, producing a reward signal that is defined at the sequence outcome level rather than at the token level. Under GRPO~\citep{shao2024deepseekmath}, a single noisy high-reward rollout propagates its advantage through per-token ratios; consequently, tokens that happen to diverge from $\pi_{\mathrm{old}}$ receive disproportionately large updates, even if they are not causally responsible for the binary outcome. In contrast, GSPO~\citep{zheng2025group} aggregates the ratio over the full response before clipping, ensuring that per-token fluctuations cancel out and only the geometric-mean deviation of the sequence survives. This alignment between clipping and reward granularity significantly reduces gradient variance in the presence of judge-model false positives and rule-based shortcut exploitation.

\paragraph{Role of $\beta = 0.04$ in GSPO.}
We use $\beta = 0.04$ as a conservative choice given the LLM-as-a-judge reward noise, noting that GSPO's original formulation omits KL regularization. RLVR begins from a LoRA-adapted policy without any SFT warmup on the target persona distribution. At step 0, $\pi_\theta \approx \pi_{\mathrm{ref}}$ and $\mathbb{D}_{\mathrm{KL}} \approx 0$, so the choice of $\beta$ governs how much cumulative drift is tolerated once advantages become informative. We found that $\beta = 0$ (unconstrained) permits rapid drift, which in preliminary runs led to format collapse and over-assertive answering on unanswerable prompts. In contrast, $\beta = 0.04$ preserved the group-relative advantage signal while bounding drift to a trust-region regime.

\paragraph{Variance-Aware Rollout Resampling.}
Group-relative policy optimization requires within-group reward variation to produce an informative relative-advantage signal. For a prompt \(x\), let
\(\{r_{x,1},\ldots,r_{x,G}\}\) denote the rewards of its \(G\) sampled completions. We classify the rollout group as \emph{live} when

\begin{equation}
\operatorname{Std}\big(r_{x,1},\ldots,r_{x,G}\big) > \epsilon,
\end{equation}

and as \emph{dead} otherwise. Dead groups arise when all completions receive the same reward and therefore contribute little or no group-relative learning signal.

When a dead group is encountered, we regenerate completions for the same prompt with an increased sampling temperature. We perform at most two retries, multiplying the temperature by \(1.25\) at each retry up to a maximum of \(1.6\), starting from an initial temperature of \(1.1\). Training retains the first live group obtained; groups that remain uninformative after the final retry are excluded from the group-relative loss.

This procedure is inspired by dynamic sampling in DAPO~\citep{yu2026dapo} but is adapted to the small effective prompt batch used in our setting. Directly discarding all zero-variance groups would remove many all-wrong positive-match groups, even though these groups reveal precisely the grounding failure that we seek to correct. Prompt-level resampling instead attempts to recover informative variation before excluding a group.

\begin{table}[t!]
\centering
\caption{Keyword and phrase list used for lexical abstention detection.}
\label{tab:abstain_keywords}
\small
\begin{tabular}{ll}
\toprule
\multicolumn{2}{c}{\textbf{Abstention Keywords / Target Surface Forms}} \\
\midrule
cannot determine & cannot be determined \\
not enough information & insufficient information \\
cannot answer & unable to determine \\
don't know from & do not know from \\
the provided context does not & not provided in the context \\
no information in the context & context does not contain \\
cannot identify & i cannot tell \\
\bottomrule
\end{tabular}
\end{table}

\section{Limitations and Broader Impacts}\label{appen:limitations}
\paragraph{Synthetic-Real Domain Bias in Post-Training.}
Real-world face images and voice recordings of consistent identities are difficult to obtain at the scale required for systematic SFT and RLVR training, particularly given privacy, consent, and licensing constraints. We therefore train on synthetic personas (TTS-generated voice clips and generated facial images) while reserving real images strictly for the Omni-Persona evaluation benchmark. This deliberate split preserves benchmark realism and also provides a natural testbed for studying how well post-training generalizes from synthetic personas to real-world distributions, a question we view as important for future omnimodal personalization research. 

\paragraph{Difficulty of Evaluation-Aligned In-Domain SFT Data.}
One could alternatively use an SFT dataset that closely matches the evaluation distribution. However, constructing such a dataset at scale is a significant challenge. Generating high-quality answers without accidentally exposing test-set information is difficult, and using test-style queries during training risks benchmark contamination. Therefore, we treat our SFT results as an analysis of broad-coverage training. Developing a reliable method to create high-quality, in-domain data while maintaining benchmark integrity remains an important goal for future research.

\paragraph{Scope of Reward Design Ablations.}
Our RLVR pipeline includes several stability filters, such as 4-gram repetition, character diversity, and sentence-repetition checks, which effectively prevent severe degeneration patterns that can occur during RL training on thinking-style models. Fine-grained reward shaping for balancing grounding and abstention is a distinct and complementary direction; we leave such ablations to future work. Importantly, we view the answer-inducing abstention collapse surfaced by our $\mathrm{TA}$ and Cal metrics as a diagnostic strength of Omni-Persona, rather than as evidence that RLVR is inherently prone to this behavior, and we expect it to be addressable with more targeted reward design.


\paragraph{Limitations of Lexical Abstention Detection.}
\label{sec:limitation_abstention}
To determine whether a model abstains on unanswerable queries, we use a simple lexical matching rule. A response is classified as an abstention if it contains any predefined phrase listed in Table~\ref{tab:abstain_keywords}. The same rule is applied both during RLVR training, where it defines the abstention-related reward signal, and at evaluation time, making abstention evaluation directly comparable without requiring an additional judge model. This lexical design may undercount valid abstentions, particularly when base or SFT models use phrases outside the predefined keyword list. Although the evaluation prompt in Table~\ref{template:memory_qa_system_prompt} instructs models to use ``I cannot determine that from the provided context'' for unsupported answers, models may still produce paraphrases or partial uncertainty statements that are not captured by the detector.

\paragraph{Privacy and Consent.}
Omni-Persona is constructed from publicly available or synthetically generated user signals, including face images, voice samples, and biographical text, and does not include personally identifiable information from real individuals without consent. However, systems built for real-user personalization must obtain explicit consent for collecting, storing, and processing biometric signals such as voice and facial imagery.

\paragraph{Limited Coverage of Perceptual Uncertainty.} The benchmark evaluates relatively fine-grained visual and auditory discrimination within a provided query--context setting; it does not fully represent open-ended uncertainty in unconstrained real-world environments. The instances nevertheless include several realistic sources of difficulty, including background speech, conversational voices, perceptually similar distractors, and missing textual information. These factors are not separately annotated according to a controlled difficulty taxonomy such as ambiguous evidence, noisy audio, occlusion, partial evidence, or highly similar candidates. The current benchmark therefore cannot quantify the degradation attributable to each source of perceptual uncertainty. A controlled analysis that isolates these uncertainty types is an important direction for future work.

\paragraph{Bias and Representation.}
Synthetic persona generation may underrepresent certain demographic groups, leading to uneven personalization quality across users. While our benchmark includes hard distractors and retrieval noise to test robustness, fairness across multilingual voice, facial, and linguistic attributes remains an important direction for future evaluation.

\paragraph{Future Directions.}
Our work remains limited to the post-retrieval grounding and generation stage. A natural next step is therefore to integrate Omni-Persona with end-to-end retrieval-augmented pipelines that jointly evaluate retrieval quality, multimodal grounding, generation, and calibration, while carefully distinguishing retriever errors from downstream grounding failures. This framework could further be extended to on-device personalization benchmarks constructed from unstructured omnimodal user signals, such as daily voice memos, image galleries, and textual interaction histories, as well as to agentic tool-use settings in which models perform complex cross-modal actions grounded in locally stored user context.



\begin{figure}[t!]
  \centering
  \includegraphics[width=\columnwidth]{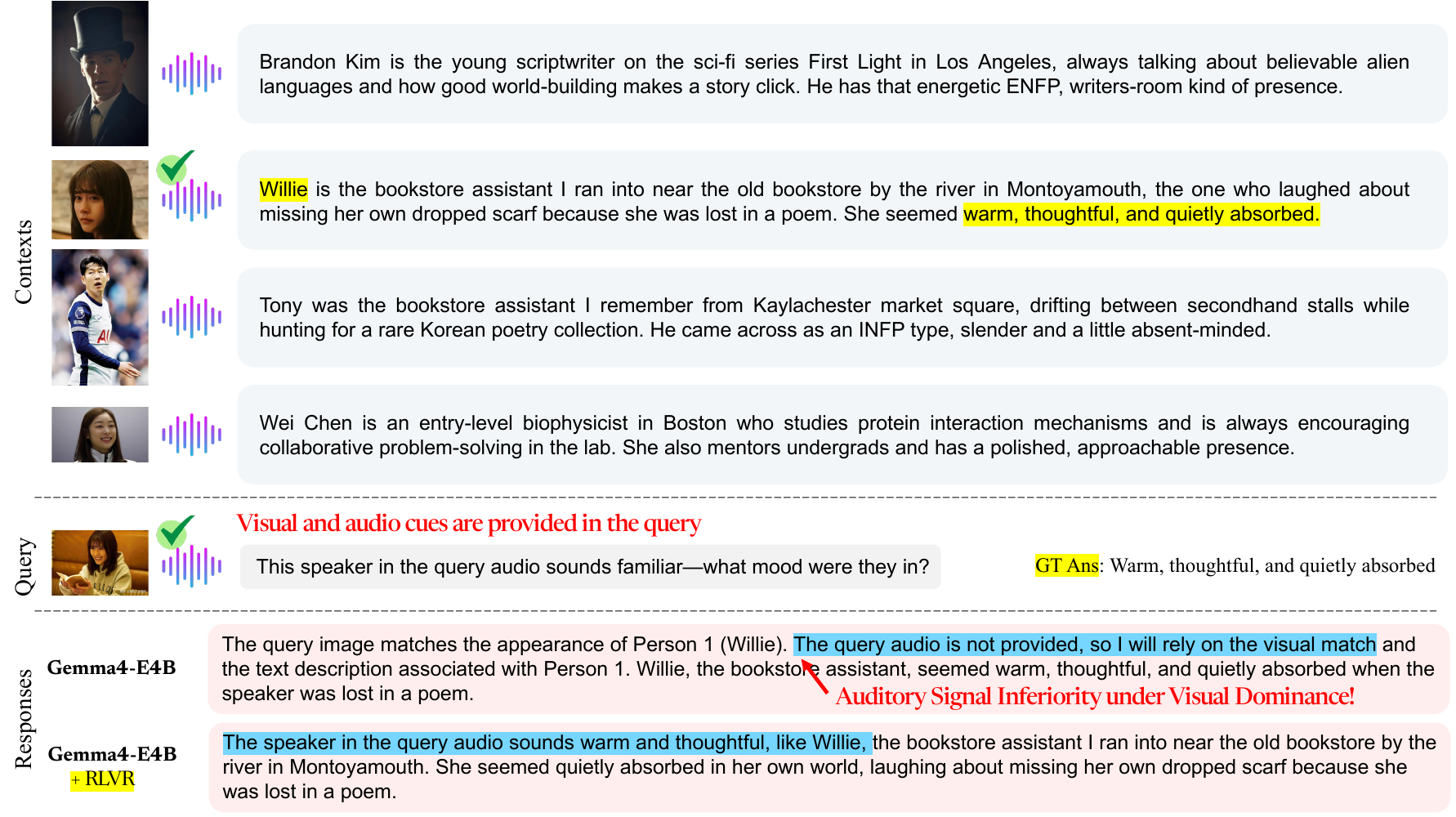}
    \caption{\textbf{Qualitative comparison of model predictions given the query and its GT answer.} When the query includes both image and audio modalities and audio is the primary perceptual cue, Gemma4-E4B tends to overlook the audio signal and rely heavily on visual cues. In contrast, the RLVR-trained model grounds its response in the relevant audio signal.}  \label{fig:qualitative}

    \vspace{-1em}
\end{figure}

\begin{table}[t!]
\centering
\footnotesize

\setlength{\aboverulesep}{0pt}
\setlength{\belowrulesep}{0pt}
\setlength{\tabcolsep}{5pt}
\renewcommand{\arraystretch}{1.05} 

\caption{\textbf{Complementary lexical and semantic metric results.} Note: BS = BERTScore, 1-FA = 1-FalseAbs, TA = TrueAbs, MLen = Mean Length.}
\label{tab:main_results_refined}

\resizebox{\textwidth}{!}{%
\begin{tabular}{l ccc cccc c | ccc | c}
\toprule

\multirow{2}{*}{\textbf{Model}} 
& \multicolumn{3}{c}{\textbf{Accuracy}} 
& \multicolumn{4}{c}{\textbf{Generation Quality}} 
& \multirow{2}{*}{\textbf{Avg}} 
& \multicolumn{3}{c|}{\textbf{Add. Metrics ($\times$100)}} 
& \multirow{2}{*}{\textbf{MLen}} \\
\cmidrule(lr){2-4} \cmidrule(lr){5-8} \cmidrule(lr){10-12}
& \textbf{Overall(Cal)} & \textbf{Ans} & \textbf{Unans} 
& \textbf{Abs-F1} & \textbf{ROUGE-L} & \textbf{Tok-F1} & \textbf{BS} 
& & \textbf{1-FA} & \textbf{TA} & \textbf{Avg} & \\
\midrule

\rowcolor{gray!15} \multicolumn{13}{l}{\textit{Closed-source Models}} \\
Gemini-3.6-Flash & \textbf{35.7} & \textbf{60.1} & \textbf{9.2} & \textbf{16.8} & \textbf{62.1} & \textbf{17.2} & \textbf{84.7} & \textbf{40.8} & 99.7 & \textbf{9.2} & \textbf{54.5} & 35.6 \\
Gemini-3.6-Flash (Think) & 28.8 & 51.2 & 4.5 & 8.5 & 55.5 & 15.8 & 84.3 & 35.5 & \textbf{100.0} & 4.5 & 52.2 & 32.5 \\
Gemini-3.5-Flash-lite & 29.7 & 51.4 & 6.1 & 11.5 & 59.2 & 16.2 & 84.7 & 37.0 & 99.5 & 6.1 & 52.8 & \textbf{40.2} \\[0.2em]
\rowcolor{gray!15} \multicolumn{13}{l}{\textit{Open-source Models}} \\
Phi-4 Multimodal & 33.2 & 40.7 & 25.1 & 37.4 & 51.9 & 12.5 & 83.8 & 40.7 & \textbf{91.8} & 25.1 & 58.5 & 77.2 \\
MiniCPM-o 4.5 (Think) & \textbf{39.3} & \textbf{41.7} & \textbf{36.8} & \textbf{48.9} & \textbf{67.6} & 5.4 & 79.8 & \textbf{45.6} & 87.5 & \textbf{36.8} & \textbf{62.1} & \textbf{716.0} \\
Interactive-Omni & 29.6 & 32.7 & 26.2 & 36.7 & 35.2 & \textbf{14.3} & \textbf{85.1} & 37.1 & 84.9 & 26.2 & 55.6 & 20.3 \\
Qwen3-Omni-30B & 33.7 & 51.2 & 14.8 & 24.4 & 58.0 & 13.2 & 84.2 & 39.9 & 94.4 & 14.8 & 54.6 & 59.0 \\

\textbf{Qwen2.5-Omni-3B} & 36.7 & 34.0 & 39.6 & 47.4 & 35.6 & 15.3 & 85.3 & 42.0 & 74.9 & 39.6 & 57.2 & 18.1 \\
\rowcolor{cyan!12} \quad SFT (1K)             & 36.4 & 33.5 & 39.6 & 47.4 & 35.6 & 15.3 & 85.3 & 41.9 & 74.9 & 39.6 & 57.2 & 18.1 \\
\rowcolor{cyan!12} \quad SFT (10K)            & \textbf{36.9} & 34.0 & \textbf{40.1} & \textbf{48.1} & 35.5 & 15.4 & 85.3 & \textbf{42.2} & 75.4 & \textbf{40.1} & \textbf{57.8} & 18.1 \\
\rowcolor{cyan!12} \quad RLVR     & 34.1 & \textbf{40.9} & 26.7 & 37.3 & \textbf{44.8} & \textbf{19.2} & \textbf{85.9} & 41.3 & \textbf{84.7} & 26.7 & 55.7 & \textbf{19.6} \\[0.2em]

\textbf{Qwen2.5-Omni-7B} & 30.9 & 38.6 & \textbf{22.6} & \textbf{31.8} & 45.4 & 14.6 & 84.8 & 38.4 & 82.1 & \textbf{22.6} & 52.3 & 35.1 \\
\rowcolor{cyan!12} \quad SFT (1K)             & 31.5 & 39.6 & \textbf{22.6} & \textbf{31.8} & 45.4 & 14.6 & 84.8 & 38.6 & 82.1 & \textbf{22.6} & 52.3 & 35.1 \\
\rowcolor{cyan!12} \quad SFT (10K)            & \textbf{31.9} & 42.7 & 20.1 & 29.2 & 47.4 & 14.7 & 84.9 & \textbf{38.7} & 84.1 & 20.1 & 52.1 & 35.8 \\
\rowcolor{cyan!12} \quad RLVR     & 28.4 & \textbf{47.8} & 7.2 & 13.2 & \textbf{55.4} & \textbf{17.8} & \textbf{85.5} & 36.5 & \textbf{98.0} & 7.2 & \textbf{52.6} & \textbf{36.7} \\[0.2em]

\textbf{Gemma4-E2B} & \textbf{32.1} & 41.7 & \textbf{21.7} & \textbf{32.4} & 52.9 & 9.9 & \textbf{80.4} & \textbf{38.7} & 88.7 & \textbf{21.7} & \textbf{55.2} & 72.8 \\
\rowcolor{cyan!12} \quad SFT (1K)             & 31.9 & 41.2 & \textbf{21.7} & \textbf{32.4} & 52.9 & 9.9 & \textbf{80.4} & 38.6 & 88.7 & \textbf{21.7} & \textbf{55.2} & 72.8 \\
\rowcolor{cyan!12} \quad SFT (10K)            & 31.3 & 40.7 & 21.2 & 31.8 & 52.2 & 9.9 & 80.3 & 38.2 & 89.0 & 21.2 & 55.1 & 72.7 \\
\rowcolor{cyan!12} \quad RLVR     & 28.1 & \textbf{44.5} & 10.3 & 19.0 & \textbf{54.9} & \textbf{10.3} & 80.3 & 35.3 & \textbf{96.7} & 10.3 & 53.5 & \textbf{76.9} \\[0.2em]

\textbf{Gemma4-E4B} & 31.2 & 46.8 & 14.2 & 23.3 & 56.5 & 15.2 & 84.2 & 38.8 & 92.8 & 14.2 & \textbf{53.5} & 45.1 \\
\rowcolor{cyan!12} \quad SFT (1K)             & 26.9 & 43.5 & 8.9 & 15.9 & 51.4 & 15.7 & \textbf{84.7} & 35.3 & \textbf{96.9} & 8.9 & 52.9 & 34.2 \\
\rowcolor{cyan!12} \quad SFT (10K)            & \textbf{33.7} & 46.8 & \textbf{19.5} & \textbf{28.9} & \textbf{58.4} & 15.2 & 83.6 & \textbf{40.9} & 85.7 & \textbf{19.5} & 52.6 & \textbf{57.5} \\
\rowcolor{cyan!12} \quad RLVR     & 30.4 & \textbf{48.6} & 10.6 & 22.4 & 58.4 & \textbf{16.1} & 84.6 & 38.7 & 94.4 & 10.6 & 52.5 & 42.9 \\[0.2em]
\bottomrule
\end{tabular}%
}
\end{table}

\section{Further Analysis on Lexical Metrics}

\label{appen:metric_analysis}

\paragraph{Lexical Metrics Validate, but Cannot Replace, the LLM Judge.}

We first examine whether our LLM-as-a-judge verdicts are consistent with reference-based lexical signals. For each model, we partition answerable predictions ($N=391$) by judge verdict and recompute lexical scores within each partition. Across all evaluated backbones, predictions judged as \textsc{Correct} consistently achieve substantially higher ROUGE-L than those judged as \textsc{Wrong}, typically by a factor of 2--4. This pattern suggests that the judge is aligned with the same reference-overlap signal captured by conventional lexical metrics. However, lexical overlap remains too sparse for free-form personalized QA. Per-item lexical overlap remains sparse: threshold-based accuracy using ROUGE-L$\,\ge\!0.5$ stays below 14\% across models. Thus, lexical metrics provide a useful sanity check for judge reliability, but they cannot serve as the primary evaluation criterion. This motivates a calibrated, behavior-aware metric that accounts for both grounded answering and appropriate abstention.



\paragraph{Post-training Regimes Show Different Behavioral Fingerprints.}

From this calibrated perspective, SFT and RLVR produce qualitatively different behavioral shifts. SFT-10K largely preserves the base-model behavior: ROUGE-L changes only marginally, and the genuine hallucination rate remains nearly unchanged. This suggests that broader supervised data alone does not reliably reshape the model's abstention boundary in open-ended personalization settings. In contrast, RLVR more directly reshapes this boundary, but in the answer-inducing direction: it shifts every backbone toward attempting answers, reducing false abstention on answerable queries, while unsupported answers on unanswerable queries rise and Cal fails to improve over the base model. This reflects a clear trade-off under our current reward design: outcome-level rewards targeting grounding and reasoning yield large answerable-recall gains for the Qwen backbones (with Gemma recall near base), but fail to induce calibrated abstention. The evaluated models illustrate the same tension from another angle. \texttt{MiniCPM-o 4.5} achieves strong calibrated behavior, whereas \texttt{Gemini-3.6-Flash} tends toward over-answering. These extremes confirm that single-axis metrics are insufficient: lexical quality, recall, hallucination, and abstention behavior can move independently. Calibrated Accuracy is therefore an appropriate headline metric for personalized omnimodal QA because it jointly penalizes blanket abstention and unsupported guessing.

\clearpage
\section{Detailed Task Taxonomy and Granularities}
\label{appendix:data_construction}
\begin{table}[h!]
\centering
\footnotesize
\renewcommand{\arraystretch}{1.05}
\setlength{\tabcolsep}{4pt}
\caption{Fine-grained task taxonomy of the Omni-Persona benchmark: 4 matching scenario groups and 18 sub-tasks derived from the PMG formulation. Each group represents a primary matching scenario defined by its query and context modalities; sub-tasks within a group sweep the retrieval target modality. Symbols: $I$ (image), $T$ (text), $A^v$ (audio/voice sample). Every sub-task additionally has an absent-persona (no-GT, $e_{q \to j}{=}0$, target=\texttt{None}) variant that requires structured abstention.}
\label{tab:task_taxonomy}
\begin{tabularx}{\linewidth}{@{} c c c c p{2.2cm} >{\raggedright\arraybackslash}X @{}}
\toprule
\textbf{ID} & \textbf{Query} & \textbf{Context} & \textbf{Target} & \textbf{Task} & \textbf{Sample question} \\
\midrule
\multicolumn{6}{@{}l}{\textit{Group 1 (I2I): Visual Identification (Query: $I$, Context: $I$)}} \\
\midrule
1-a & $I$ & $I$ & $T$              & Biography         & \textit{Who is this person, and what is their job?} \\
1-b & $I$ & $I$ & $A^v$ & Dialogue recall   & \textit{Looking at this person in the image, I'm trying to remember what they said before.} \\
1-c & $I$ & $I$ & $I$              & Appearance        & \textit{This person in the image seems familiar. What did they look like?} \\
1-d & $I$ & $I$ & $A^v$ & Emotion           & \textit{When I met this person in the image, what mood were they in?} \\
1-e & $I$ & $I$ & $A^v$ & Environment       & \textit{Who is this person in the image, and where were we?} \\
\midrule
\multicolumn{6}{@{}l}{\textit{Group 2 (A2A): Voice Identification (Query: $A^v$, Context: $A^v$)}} \\
\midrule
2-a & $A^v$ & $A^v$ & $T$              & Biography         & \textit{Who is the speaker in the query audio, and what is their affiliation?} \\
2-b & $A^v$ & $A^v$ & $A^v$ & Dialogue recall   & \textit{Whose voice is in the query audio, and what had they said earlier?} \\
2-c & $A^v$ & $A^v$ & $I$              & Appearance & \textit{I recognize this speaker from the audio. Can you describe their appearance?} \\
2-d & $A^v$ & $A^v$ & $A^v$ & Emotion           & \textit{This speaker in the query audio sounds familiar. What mood were they in?} \\
2-e & $A^v$ & $A^v$ & $A^v$ & Environment       & \textit{This speaker in the query audio belongs to someone I know. What was happening around them?} \\
\midrule
\multicolumn{6}{@{}l}{\textit{Group 3 (T2T): Same-Modal Semantic (Query: $T$, Context: $T$)}} \\
\midrule
3-a & $T$ & $T$ & $T$              & Biography         & \textit{The professional esports player, what was their occupation?} \\
3-b & $T$ & $T$ & $I$              & Appearance        & \textit{I'm thinking of the bookstore café barista. What were they wearing?} \\
3-c & $T$ & $T$ & $T$ & Emotion           & \textit{The freelance illustrator, how were they feeling?} \\
3-d & $T$ & $T$ & $T$  & Environment       & \textit{The gallery assistant, where were we when that happened?} \\
\midrule
\multicolumn{6}{@{}l}{\textit{Group 4 (T2Any): Cross-Modal Semantic (Query: $T$, Context: $A^v$)}} \\
\midrule
4-a & $T$ & $T$ & $T$              & Biography         & \textit{The one who said that, what was their affiliation?} \\
4-b & $T$ & $T$ & $I$              & Appearance & \textit{The person who talked about a peaceful reunion over poetry and a dropped scarf, what did they look like?} \\
4-c & $T$ & $T$ & $A^v$ & Emotion           & \textit{The person who talked about an unexpected rainy-day encounter outside an art gallery, how were they feeling then?} \\
4-d & $T$ & $T$  &$T$  & Environment       & \textit{Can you identify the person who talked about a memorable fan encounter at a gaming convention, and tell me where we were?} \\
\bottomrule
\end{tabularx}
\end{table}

\clearpage
\begin{figure}[t!]
  \centering
\includegraphics[width=0.95\columnwidth]{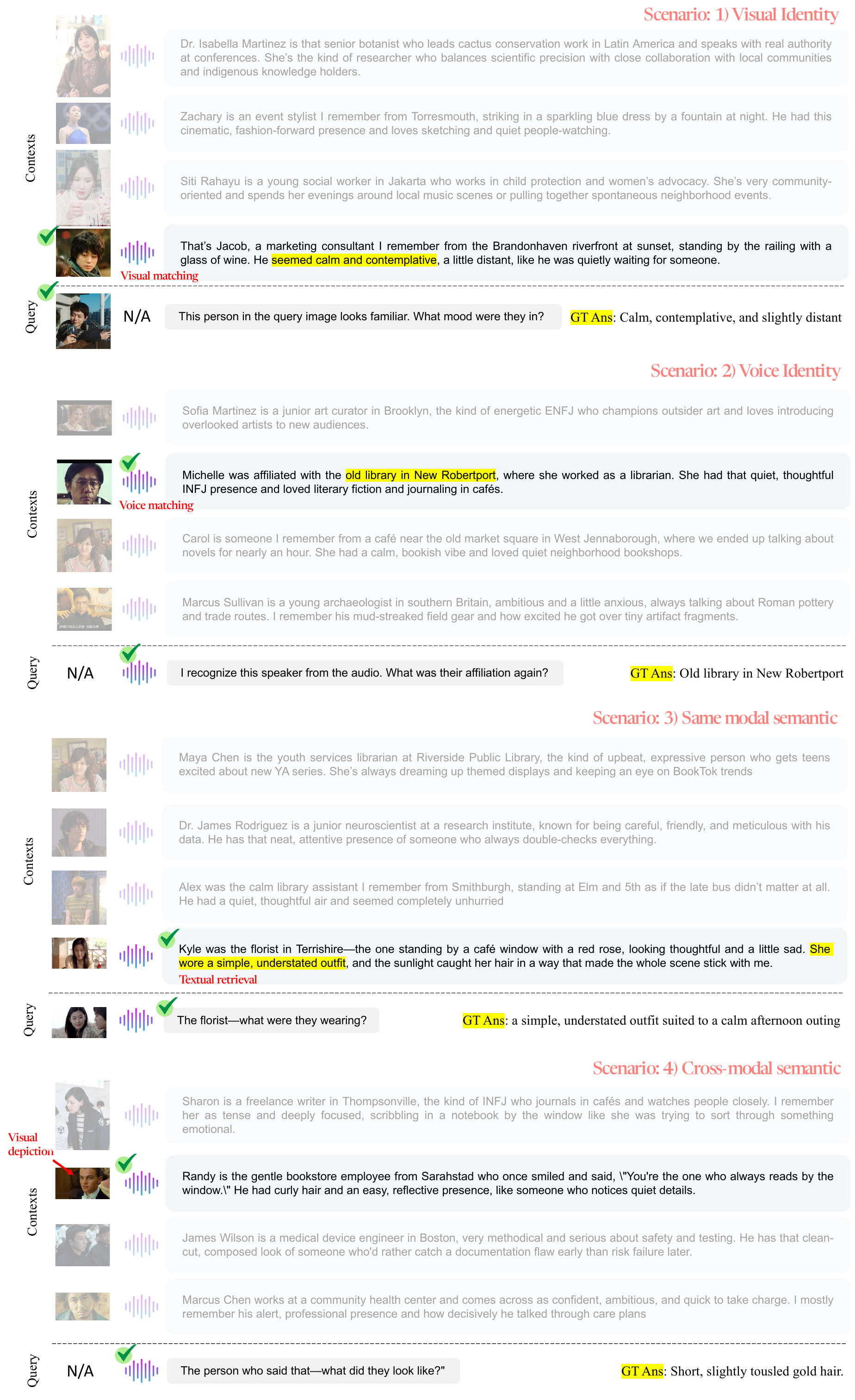}
\caption{Illustrative examples for each scenario within the Omni-Persona benchmark. Specifically, only answerable instances (where the target persona is present) are visualized in this figure.} \label{fig:image_scenario_all}
\end{figure}

\clearpage
\section{Used Templates for Dataset Construction}\label{appendix:template_construction}

\begin{table}[h!]
\centering
\captionsetup{justification=centering}
\caption{Fixed prompt template used for evaluation in our Omni-Persona benchmark.}
\vspace{-0.5em}
\begin{tcolorbox}[colframe=black!50!white, colback=white,
  boxrule=0.4pt, arc=6pt, width=1.0\textwidth,
  left=6pt, right=6pt, top=6pt, bottom=6pt]

\textbf{User Prompt Prompt:} \\
You are a personal memory assistant for a multimodal personalization benchmark.
\newline

[Context Setup] \\
You will receive memory contexts about four different people (\texttt{Context 0} through \texttt{Context 3}). Each context may contain some combination of:
\begin{itemize}[leftmargin=1.5em]
    \item A photo of the person (\texttt{image})
    \item A voice recording or spoken clip of the person (\texttt{audio})
    \item A text description with personal details such as name, job, hobbies, personality, location, etc.
\end{itemize}

After the four memory contexts, you will receive a \texttt{Query} section that may include:
\begin{itemize}[leftmargin=1.5em]
    \item A query image showing a person (use it to identify which context person this is)
    \item A query audio clip of a person speaking (use it to identify which context person this is)
    \item A question asking for specific information about that person
\end{itemize}

[Task] \\
\begin{enumerate}[leftmargin=1.5em]
    \item Carefully examine all four contexts, attending to every available modality.
    \item Use the query image, audio, or text to identify which context (\texttt{Person 0--3}) the query refers to.
    \begin{itemize}[leftmargin=1.5em]
        \item If the query includes an image, match the face or appearance to the context images.
        \item If the query includes an audio clip, match the voice to the context audio recordings.
        \item If the query is text-only, use the semantic description to identify the correct person.
    \end{itemize}
    \item Answer the question using only the information from the matched context.
\end{enumerate}

[Response Rules] \\
\begin{enumerate}[leftmargin=1.5em]
    \item Before answering, briefly reason through which person matches the query and what relevant information their context contains.
    \item Provide a specific and informative answer in \textbf{1--3 sentences}, including all relevant details available in the matched context.
    \item Base the answer solely on the provided contexts; do not hallucinate or infer beyond what is explicitly given.
    \item Do not repeat the question.
    \item Use \texttt{\textcolor{red}{I cannot determine that from the provided context.}}  only as a last resort, when the requested information is genuinely absent from all modalities of the matched person's context.
    \item If any modality provides a partial or indirect answer, use it rather than abstaining.
\end{enumerate}

[Output Behavior] \\
The response should first briefly identify the matched person and summarize the relevant evidence, and then provide the final answer grounded only in the corresponding context.

\end{tcolorbox}
\label{template:memory_qa_system_prompt}
\end{table}
\begin{table}[h!]
\centering
\captionsetup{justification=centering}
\caption{Judge prompt used for answer correctness.}
\vspace{-0.5em}
\begin{tcolorbox}[colframe=black!50!white, colback=white,
  boxrule=0.4pt, arc=6pt, width=1.0\textwidth,
  left=6pt, right=6pt, top=6pt, bottom=6pt]

\textbf{Evaluation Judge Prompt:} \\
You are an evaluation judge for a question-answering benchmark. You will be given a gold answer and a model prediction, and must decide whether the prediction is correct.
\newline

[Judging Rules] \\
\begin{enumerate}[leftmargin=1.5em]
    \item For factual answers (\textit{non-abstain}), the prediction is \texttt{CORRECT} if its final answer conveys the same core information as the gold answer, even if the wording differs.
    \begin{itemize}[leftmargin=1.5em]
        \item Minor paraphrasing is acceptable.
        \item The prediction is also \texttt{CORRECT} if the gold answer is explicitly contained in the prediction as a clear and stated fact.
    \end{itemize}

    \item The model receives multiple memory contexts and must identify the correct person before answering.
    \begin{itemize}[leftmargin=1.5em]
        \item If the prediction lists multiple conflicting answers across different people, it is \texttt{WRONG}.
        \item If the gold answer is only mentioned in passing while a different answer is presented as the model’s conclusion, it is \texttt{WRONG}.
        \item If the model hedges without committing to a final answer, it is \texttt{WRONG}.
    \end{itemize}

    \item For abstain gold answers (\texttt{``I cannot determine...''}), the prediction is \texttt{CORRECT} only if it also abstains.
    \begin{itemize}[leftmargin=1.5em]
        \item Any concrete answer in this case is \texttt{WRONG}.
    \end{itemize}

    \item Output exactly one word: \texttt{CORRECT} or \texttt{WRONG}.
    \begin{itemize}[leftmargin=1.5em]
        \item Do not provide any explanation.
    \end{itemize}
\end{enumerate}

[User Input Template] \\
\vspace{0.25em}
\noindent\textbf{Input:}\\
\texttt{Gold answer: \{gold\}}\\
\texttt{Model prediction: \{pred\}}\\
\texttt{Verdict:}

\end{tcolorbox}
\label{template:judge_prompt}
\end{table}
\begin{table}[h!]
\centering
\captionsetup{justification=centering}
\caption{Visualization of the structured prompt used for automated persona profiling and attribute enrichment.}
\vspace{-0.5em}
\begin{tcolorbox}[colframe=black!50!white, colback=white,
  boxrule=0.4pt, arc=6pt, width=1.0\textwidth,
  left=6pt, right=6pt, top=6pt, bottom=6pt]

\textbf{Personal Profile Builder Prompt (JSON-only):}
\newline
You are a personal profile builder for a multimodal personalization benchmark.
\newline
\newline
[Task] \\
Your job has two stages:
\begin{enumerate}[leftmargin=1.5em]
    \item \textbf{EXTRACT}: pull out any facts explicitly stated in the source text (e.g., name, job, MBTI).
    \item \textbf{ENRICH}: for every field still \texttt{null} after extraction, invent a plausible and realistic value consistent with the dialogue style.
\end{enumerate}

Think like a person writing down memory notes: assign believable personal attributes such as a hobby, a past travel experience, a job, an MBTI type, a memorable quote, an emotional state, or a physical appearance so that the profile feels like a real person someone might remember.
\newline
\newline

[Constraints]
\begin{enumerate}[leftmargin=1.5em]
    \item Never contradict explicitly stated facts in the source text.
    \item Keep each value concise (1--2 sentences). Values for experience/preference fields should reflect realistic, personal anecdotes.
    \item Output must be strict JSON matching the profile schema.
\end{enumerate}

[Input] \\
\texttt{concept\_id: \{concept\_id\}} \quad \texttt{source\_type: \{dialogue\_type\}} \quad \texttt{SOURCE: \{source\_text\}}
\newline
\newline
[Profile Schema]
\begin{verbatim}
{
  "hobby": "<string>",
  "major": "<string|null>",
  "affiliation": "<string|null>",
  "role": "<string>",
  "interest": "<string>",
  "topic": "<string>",
  "quote": "<string>",
  "location": "<string>",
  "environment": "<string>",
  "emotion": "<string>",
  "appearance": "<string>",
  "clothing": "<string>",
  "hairstyle": "<string>",
  "voice_identity": "<string|null>",
  "source_style": "<string>",
  "travel_experience": "<string|null>",
  "food_preference": "<string|null>",
  "life_event": "<string|null>"
}
\end{verbatim}

\end{tcolorbox}
\label{template:profile_builder_prompt}
\end{table}

\clearpage



\end{document}